\documentclass[journal]{IEEEtran}

\usepackage{amsthm,amsmath,amsfonts,bbm}
\usepackage{array}
\usepackage{textcomp}
\usepackage{stfloats}
\usepackage{url}
\usepackage{verbatim}
\usepackage{graphicx}
\usepackage{cite}

\usepackage{amssymb}
\usepackage{bm}
\usepackage{multirow}
\usepackage{booktabs}
\usepackage{lipsum}
\usepackage{enumerate}
\usepackage{multicol}
\usepackage{multirow}
\usepackage{subfigure}
\usepackage{stfloats}
\usepackage{bbding}

\usepackage{cuted}
\usepackage{capt-of}
\usepackage{xcolor}

\usepackage{algorithm}
\usepackage{algorithmic}

\usepackage[colorlinks,
linkcolor=blue,
anchorcolor=blue,
citecolor=blue]{hyperref}

\usepackage{halloweenmath}
\allowdisplaybreaks[4]

\newtheorem{lemma}{Lemma}

\newtheorem{definition}{Definition}
\newtheorem{proposition}{Proposition}

\newtheorem{theorem}{Theorem}

\newtheorem{assumption}{Assumption}
\newtheorem{corollary}{Corollary}

\begin{document}

\title{Distributed Dynamic Associative Memory via Online Convex Optimization}

\author{Bowen Wang,
Matteo Zecchin,~\IEEEmembership{Member,~IEEE}, and Osvaldo Simeone,~\IEEEmembership{Fellow,~IEEE}
\vspace{-8em}
\thanks{An earlier version of this paper \cite{wang2025distributed} was presented at the 2026 IEEE International Conference on Acoustics, Speech, and Signal Processing (ICASSP)  [DOI: 10.1109/ICASSP55912.2026.11464435].}
\thanks{This work was partially supported  by the Open Fellowships of the EPSRC (EP/W024101/1)  and by the EPSRC project (EP/X011852/1).}  
\thanks{
Bowen Wang is with the Department of Engineering, King’s College London, London WC2R 2LS, U.K. (e-mail: bowen.wang@kcl.ac.uk).}
\thanks{Matteo Zecchin is with the Communication Systems Department, EURECOM, 06904 Sophia Antipolis, France (e-mail: zecchin@eurecom.fr).}
\thanks{Osvaldo Simeone is with the Institute for Intelligent Networked Systems (INSI) at Northeastern University London, London E1 8PH, U.K. (e-mail: o.simeone@northeastern.edu).}
}

\maketitle
\begin{strip}
	
	\centering
	\begin{minipage}{0.34\textwidth}
		\centering
		\includegraphics[width=\linewidth]{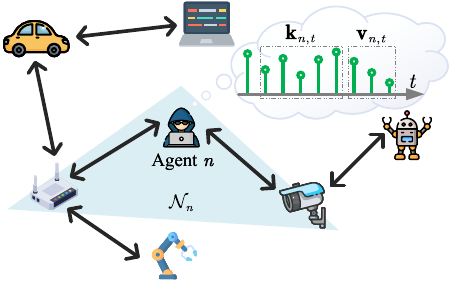}\\[0.5ex]
		\small (a) Physical connectivity
	\end{minipage}
	\hspace{6em}
	\begin{minipage}{0.34\textwidth}
		\centering
		\includegraphics[width=\linewidth]{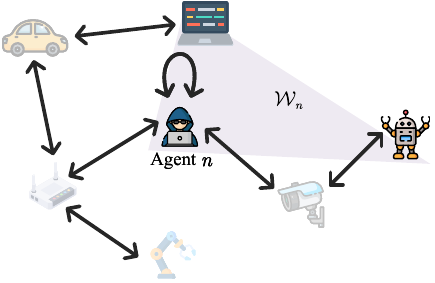}\\[0.5ex]
		\small (b) Logical connectivity
	\end{minipage}
	
	\captionof{figure}{(a) In a \textit{distributed dynamic associative memory} (DDAM)  system, each agent maintains a local memory, and corresponding AM mechanism, by processing local streaming data, as well as by interacting with other agents over a physical network. 
	Each agent $n$ collects streaming data in the form of key $\mathbf{k}_{n,t}$ and value $\mathbf{v}_{n,t}$ over discrete time $t=1,2,...,T$. 
	The blue region  represents the subset $\mathcal{N}_n$ of physical  neighbors  that can directly exchange information with agent $n$ over the  network. 
	(b) The goal of DDAM is for the memory at each agent not only to recall associations from its own data, but also from a subset of other agents.  
	The purple region in the figure represents the logical subset $\mathcal{W}_n$ of agents whose data is relevant for agent $n$. 
	The nodes in subset $\mathcal{W}_n$ may not be physical neighbors of agent $n$. 
	Through inter-agent communication over the physical network, DDAM aims at recalling information from both local and logically related agents in an online fashion.}
	\label{fig:1}
\end{strip}

\vspace{-2em}
\begin{abstract}
	An associative memory (AM) enables cue–response recall, and it has recently been recognized as a key mechanism underlying modern neural architectures such as Transformers. In this work, we introduce the concept of distributed dynamic associative memory (DDAM), which extends classical AM to settings with multiple agents and time-varying data streams. In DDAM, each agent maintains a local AM that must not only store its own associations but also selectively memorize information from other agents based on a specified interest matrix. To address this problem, we propose a novel tree-based distributed online gradient descent algorithm, termed DDAM-TOGD, which enables each agent to update its memory on the fly via inter-agent communication over designated routing trees. We derive rigorous performance guarantees for DDAM-TOGD, proving sublinear static regret in stationary environments and a path-length dependent dynamic regret bound in non-stationary environments. These theoretical results provide insights into how communication delays and network structure impact performance. Building on the regret analysis, we further introduce a combinatorial tree design strategy that optimizes the routing trees to minimize communication delays, thereby improving regret bounds. Numerical experiments demonstrate that the proposed DDAM-TOGD framework achieves superior accuracy and robustness compared to representative online learning baselines such as consensus-based distributed optimization, confirming the benefits of the proposed approach in dynamic, distributed environments.
\end{abstract}

\section{Introduction}
\subsection{Context and Motivation}

An associative memory (AM), originally rooted in psychology and cognitive science, refers to the ability to store cue–response pairs and to recall the corresponding response when a cue is presented  \cite{tulving1972episodic}. For instance, seeing an image of a banana can evoke its distinctive smell and taste; hearing a person’s name can bring to mind their appearance, habits, or personality. This principle of cue-based recall has inspired developments across neuroscience, signal processing, and machine learning.

Most recently, modern sequence models, notably the Transformer \cite{vaswani2017attention}, have been interpreted as implementing an AM mechanism \cite{behrouz2025s,wang2025test,zhong2025understanding}. The attention operation in Transformers performs content-based retrieval of information, effectively recalling relevant \textit{responses} (\textit{values}) based on presented \textit{cues} (\textit{keys}). In fact, inference in such models can be viewed as a form of online memory building and retrieval, which can be formalized via \textit{online convex optimization} (OCO) \cite{orabona2019modern,shalev2012online}.

However, existing AM mechanisms are centralized: all information is processed and stored by a single model or agent. 
In contrast, many real-world applications operate in inherently distributed environments. 
As illustrated in Fig. \ref{fig:1}, examples include sensor networks, internet-of-things (IoT) systems, and cellular wireless networks, where multiple agents, such as devices, sensors, or infrastructure nodes, observe only local data \cite{gezici2005localization,hussain2020machine,popovski20185g,rabbat2004distributed}. 
In such networks, not all other agents' data is equally useful: each agent's task depends only on a specific subset of peers whose observations are logically related to its own (see Fig. \ref{fig:1}(b)). 

As a concrete example, consider an IoT-based surveillance network in which multiple camera sensors are deployed across a large area, such as an airport terminal or an urban district. 
Each camera observes a local stream of key–value pairs, where values encode visual features extracted from incoming frames and keys encode associated metadata.
A single camera's view is often partial or occluded, so accurate recall benefits from associations captured by a subset of other cameras monitoring the same scene or tracking the same object (see Fig. \ref{fig:1}(a)).
Consequently, each camera should selectively memorize data from a specific subset of logically relevant peers.

In multi-agent networks, such as the aforementioned IoT networks,  data often arrive as a continuous, non-stationary stream \cite{hamilton2020time}. 
Over time, the underlying data distribution and target associations may evolve. 
This scenario is common in tasks including traffic analysis, in which data statistics vary as a function of the time of day and/or of the season \cite{sone2020wireless}.
A static AM may become outdated as new data arrive or as the environment changes. 
Therefore, it is important to develop a dynamic AM that can adapt online to non-stationary data, updating stored associations, retrieving new relevant information, and forgetting outdated associations as needed.

In summary, many practical settings demand an online, distributed AM  that enables multiple agents to jointly and adaptively memorize relevant information over time.
Motivated by these considerations, this paper focuses on the study of \emph{distributed dynamic associative memory} (DDAM). 
Our goal is to bridge the gap between classical centralized AM mechanisms and the requirements of distributed, time-varying environments. 
In the following, we outline the state of the art and then summarize our contributions.

\vspace{-1em}
\subsection{State of the Art}\label{sec:1_B}
\noindent \textbf{Associative Memory (AM):}
Early studies of AM trace back to the Hopfield network \cite{hopfield1982neural}, where memory retrieval is modeled as the minimization of an energy function toward attractor states representing stored patterns.
Building upon this formulation, subsequent works proposed dense \cite{krotov2016dense} and continuous \cite{krotov2020large} Hopfield networks, which generalize the energy function and improve the memory capacity.
These models \cite{hopfield1982neural,krotov2016dense,krotov2020large} constitute a family of energy-based AMs, where inference corresponds to energy minimization.

More recently, inter-token mixing in Transformers \cite{vaswani2017attention} and related sequence models has been interpreted as an AM mechanism \cite{ramsauer2020hopfield,katharopoulos2020transformers,yang2023gated,schlag2021linear,lin2025forgetting}. 
Examples include \emph{linear AM} mechanisms, such as linear attention \cite{katharopoulos2020transformers}, gated linear attention \cite{yang2023gated}, DeltaNet \cite{schlag2021linear}, and more general state-space models (SSMs) \cite{gu2023mamba,dao2024transformers}, which offer linear time inference and constant memory cost. 
Linear AM mechanisms underlie a number of recent large-scale language models, such as NVIDIA's Nemotron-H \cite{blakeman2025nemotron}, Kimi Linear \cite{team2510kimi} and Qwen3-Next \cite{qwen2025qwen3next}.
Mathematically, linear AM can be formulated as an \emph{online convex optimization} (OCO) problem \cite{orabona2019modern,shalev2012online}.
This paper adopts this formulation as the foundation for studying AM in distributed, time-varying environments.

\noindent \textbf{Online Convex Optimization (OCO):} OCO   \cite{orabona2019modern,shalev2012online} is performed over a sequence of consecutive rounds, where at each round a learner selects a decision.
After the decision is made, a cost function is revealed and the learner incurs a corresponding loss.
The learner then updates its parameters based on the observed loss to improve subsequent performance.
This procedure may model AM \cite{behrouz2025s,wang2025test,zhong2025understanding}, where cues evoke responses that are iteratively corrected through feedback.
In OCO, the performance metric is the \textit{regret} \cite{orabona2019modern,shalev2012online}, which can be categorized as either static or dynamic.

The \textit{static regret} \cite{orabona2019modern,shalev2012online,zinkevich2003online} measures the difference between the cumulative loss incurred by the algorithm and that of the best fixed decision in hindsight.
For convex losses, the optimal static regret bound is known to be of order $\mathcal{O}(\sqrt{T})$, where $T$ is the time horizon, which can be achieved by algorithms such as \textit{online gradient descent} (OGD) \cite{zinkevich2003online,hazan2016introduction}.
In contrast, the \textit{dynamic regret} \cite{zinkevich2003online,yang2016tracking,zhao2020dynamic,zhang2019adaptive} quantifies the performance gap with respect to a time-varying comparator sequence, capturing the algorithm's adaptability to non-stationary environments.
The optimal dynamic regret also scales at $\sqrt{T}$, with multiplicative factors that depend on measures of environmental variation, such as the path-length \cite{zhang2018adaptive}, the squared path-length \cite{zhang2017improved}, and the functional variation \cite{besbes2015non}.

\noindent \textbf{Distributed learning:} Distributed learning has been an active research direction for decades. 
Depending on the learning objective, it can be broadly categorized into single-task learning and multi-task learning \cite{cao2023communication}. 
Within single-task distributed learning, it can be further divided into parameter-server based and fully decentralized paradigms \cite{cao2023communication}. 
In the parameter-server based settings, the most well-known paradigm is federated learning \cite{kairouz2021advances}, where each agent optimizes a local objective using its own data and periodically aggregates model updates to obtain a global model. 
In contrast, fully decentralized learning is based on inter-agent message passing \cite{xin2020general}. 
Specifically, in consensus-based distributed learning \cite{dimakis2010gossip}, agents exchange information only with their neighbors and iteratively update to reach an agreement on a common model.
In multi-task distributed learning formulations, the problem is to learn multiple heterogeneous, but generally related tasks distributed over agents \cite{zhang2021survey}.
Existing approaches either infer task correlations from data-driven similarity measures \cite{liu2017distributed} or define them explicitly through spatial or geographical adjacency \cite{nassif2020multitask}.

None of the reviewed procedures directly addresses the DDAM problem. 
Federated learning and consensus-based methods \cite{kairouz2021advances,xin2020general,cao2023communication} aim to learn a single global model shared across agents, whereas DDAM requires each agent to maintain its own memory and to selectively incorporate information from a subset of logically relevant peers. 
Multi-task learning instead focuses on discovering or modeling relationships among heterogeneous tasks \cite{zhang2021survey,liu2017distributed,nassif2020multitask}, whereas in DDAM the logical connectivity is prescribed and must be exploited to drive selective memorization under streaming, time-varying data. 
A new paradigm is therefore needed to solve the DDAM problem, motivating this work.

\vspace{-1em}
\subsection{Main Contributions}
In this paper, we introduce the problem of DDAM and propose a novel DDAM protocol with both static and dynamic regret guarantees.
The main contributions of this paper are summarized as follows.
\begin{itemize}
	\item \textbf{Formulation of DDAM:} As illustrated in Fig. \ref{fig:1}, we formalize the DDAM problem, in which $N$ agents collaboratively optimize their local associative memories in an online fashion. 
	Each agent $n$ processes its own stream of data, and aims to memorize not only its local input–output associations but also information from a selected subset of other agents. 

	\item \textbf{Novel protocol -- DDAM-TOGD:} We propose \textit{DDAM via tree-based OGD} (DDAM-TOGD) as an efficient solution to the DDAM problem. 
	In DDAM-TOGD, each agent runs an instance of OGD that is augmented with network communication. 
	Specifically, every agent periodically broadcasts its memory parameters along a spanning tree that connects to all agents whose data it needs. 
	Those agents compute gradient feedback using their local data and send the gradients back along the reverse path. 
    
    \item \textbf{Regret analysis:} We rigorously analyze the performance of DDAM-TOGD in terms of regret. 
    Specifically,  we prove that in static environments, DDAM-TOGD achieves sublinear static regret. 
    Hence, each agent’s memory asymptotically approaches the best \textit{fixed memory} in hindsight. 
    Moreover, for dynamic environments, we prove a dynamic regret bound that scales with the path-length of the time-varying optimal memories. 
    Accordingly, if the underlying optimal association for each agent changes slowly over time (small path-length), DDAM-TOGD can track these changes closely and incur low regret.
    In contrast, if the environment changes more rapidly, the regret grows accordingly, which is inevitable in the worst case.

    \item \textbf{Communication-efficient tree design:} The regret analysis of DDAM-TOGD provides insights into how communication delays affect performance.  Building on this insight, we propose a combinatorial optimization approach to design the communication trees for each agent.

	\item \textbf{Experimental validation:} We validate the proposed DDAM framework through extensive numerical experiments. We compare DDAM-TOGD against two representative baselines: a centralized OGD oracle \cite{shalev2012online,orabona2019modern,zinkevich2003online}, which assumes each agent has access to full information, and consensus-based distributed OGD (C-DOGD) \cite{Yan2013DisOCO}, which restricts agents to local neighbor communication under a shared memory objective. The results show that DDAM-TOGD closely tracks the centralized OGD oracle and substantially outperforms C-DOGD across both stationary and dynamic settings.
\end{itemize}

\vspace{-1em}
\subsection{Organization and Notation}
The remainder of this paper is organized as follows.
Sec. \ref{sec:2} defines the problem formulation.
Sec. \ref{sec:3} presents the state-of-the-art protocols. 
Sec. \ref{sec:4} proposes the DDAM protocol and a communication-efficient tree design.
Sec. \ref{sec:5} provides a proof sketch of the regret guarantees.
Finally, Sec. \ref{sec:6} illustrates the experimental setting and results, and Sec. \ref{sec:7} concludes the paper.
For convenience, Table \ref{tab:notation} summarizes the key notations used throughout the paper.

\begin{table}[t]
	\centering
	\caption{Summary of Key Notations}
	\label{tab:notation}
	\renewcommand{\arraystretch}{1.2}
	\begin{tabular}{ll}
		\hline
		\textbf{Symbol} & \textbf{Description} \\
		\hline \hline
		\multicolumn{2}{c}{\textit{Network and agents}} \\
		$N$, $\mathcal{N}$ & Number of agents and set of agent indices \\
		$\mathcal{G} = (\mathcal{N}, \mathcal{E})$ & Physical communication graph \\
		$\mathcal{N}_n$ & Set of physical neighbors of agent $n$ \\
		$\mathbf{W}$, $w_{n,m}$ & Logical weight matrix and its entries \\
		$\mathcal{W}_n$ & Set of agents logically relevant to $n$ \\
		\hline
		\multicolumn{2}{c}{\textit{Data and memory}} \\
		$T$, $t$ & Time horizon and discrete time index \\
		$\mathbf{k}_{n,t}$, $\mathbf{v}_{n,t}$ & Key and value at agent $n$, time $t$ \\
		$d_k, d_v$ & Key and value dimensions \\
		$\mathbf{X}_{n,t}$, $\mathcal{X}$ & AM parameters of agent $n$ and its design domain \\
		\hline
		\multicolumn{2}{c}{\textit{Cost functions and regret}} \\
		$f_{m,t}(\cdot)$ & Memory retrieval cost for agent $m$ at time $t$ \\
		$\mathbf{G}^n_{m,t}$ & Gradient $\nabla f_{m,t}(\mathbf{X}_{n,t})$ \\
		$\mathbf{U}^*_n$, $\mathbf{U}_{n,t}$ & Static and dynamic comparators for agent $n$ \\
		$\text{PL}_n^T$ & Path-length of comparator sequence \\
		$\text{S-Reg}(T)$, $\text{D-Reg}(T)$ & Static and dynamic regrets over horizon $T$ \\
		\hline
		\multicolumn{2}{c}{\textit{Algorithm parameters and constants}} \\
		$\eta_n$ & Learning rate of agent $n$ \\
		$\tau_{n,m}$ & Round-trip delay between agents $n$ and $m$ \\
		$\tau_{n,\max}$, $\Delta\tau_n$, $\tau_{n,\text{sum}}$ & Max, spread, and sum of delays from agent $n$ \\
		$B$ & Diameter of AM design domain $\mathcal{X}$ \\
		$G_n$ & Gradient norm bound at agent $n$ \\
		$K_n$ & $\max_m w_{n,m} G_m$ \\
		\hline
	\end{tabular}
\end{table}

\section{Problem Statement}\label{sec:2}
We study DDAM in a setting with networked $N$ agents.
As shown in Fig. \ref{fig:1}, each agent $n$ processes a stream of \textit{keys} $\mathbf{k}_{n , t} \in \mathbb{R}^{d_k}$ and \textit{values} $\mathbf{v}_{n , t} \in \mathbb{R}^{d_v}$, which evolve over discrete time $t = 1 , 2 , ... , T$.
For example, the key $\mathbf{k}_{n,t}$ may consist of  context samples in a time series, and the value $\mathbf{v}_{n,t}$ may include future samples of the same time series.
Alternatively, key $\mathbf{k}_{n,t}$ and value $\mathbf{v}_{n,t}$ may represent embeddings of streaming tokens for multi-modal memory recall tasks \cite{behrouz2025s,wang2025test,zhong2025understanding}.
The goal is to dynamically optimize an AM mechanism at each $n$-th agent, ensuring that at each time $t$, agent $n$ not only recalls its past data $\{ \mathbf{k}_{n , t'} , \mathbf{v}_{n , t'}\}_{t' = 1}^t $, but also data $\{ \mathbf{k}_{m , t'} , \mathbf{v}_{m , t'}\}_{t' = 1}^t $ that is processed by a subset of other agents $m \ne n$.

\vspace{-1em}
\subsection{Setting}
Denote as $\mathbf{X}_{n , t} \in \mathcal{X}$ the parameters of the AM mechanism maintained at agent $n$ at time $t$, where $\mathcal{X}$ is the design domain for parameter $\mathbf{X}_{n , t}$.
For example, for a \textit{linear AM}, the parameters $\mathbf{X}_{n , t}$ correspond to a $d_v \times d_k$ matrix, and the AM mechanism returns the estimated value $\mathbf{X}_{n , t} \mathbf{k}_{n , t}$ for the input key $\mathbf{k}_{n , t}$.
More generally, a linear AM mechanism may operate in a $d_{k'}$-dimensional space via a $d_v \times d_{k'}$ matrix $\mathbf{X}_{n,t}$ returning the output vector $\mathbf{X}_{n , t} \phi(\mathbf{k}_{n , t})$, where $\phi(\cdot) \in \mathbb{R}^{d_{k'}}$ is a non-linear feature-extraction function.

The AM mechanism at agent $n$ can be generally queried to recall data $( \mathbf{k}_{m,t}, \mathbf{v}_{m,t} )$ of any agent $m$, including itself.
The performance of AM $\mathbf{X}_{n,t}$ at agent $n$ in recalling the data $(\mathbf{k}_{m,t}, \mathbf{v}_{m,t})$ of agent $m$ is measured by a cost function $f_{m,t} (\mathbf{X}_{n,t})$.
Commonly used memory retrieval cost functions for AM are of the form
\begin{equation}\label{eq:cost_func}
	f_{m , t} ( \mathbf{X} ) = \ell ( \mathbf{X} , \mathbf{k}_{m , t} , \mathbf{v}_{m , t} ) + \mathcal{R}(\mathbf{X}) , 
\end{equation}
where $\ell ( \mathbf{X}  , \mathbf{k}_{m , t} , \mathbf{v}_{m , t} )$ is a loss function comparing the true value $\mathbf{v}_{m,t}$ with the value estimated based on memory $\mathbf{X}$, and $\mathcal{R} (\mathbf{X})$ is a regularization function.
Examples of cost functions for linear AM are summarized in Table~\ref{tab:loss}.

\begin{table}[t]
	\centering
	\setlength{\tabcolsep}{2pt}
	\caption{Memory retrieval cost functions \(f(\mathbf{X})\) of the form \eqref{eq:cost_func} for different variants of Linear Attention \cite{katharopoulos2020transformers}, where $\mathbf{X}$ represents the memory matrix, $\mathbf{k}$ is the key vector, and $\mathbf{v}$ is the corresponding value vector. ($\bm{\psi}$ is a gating vector with binary entries that is independent of $\mathbf{X}$, and $\phi(\cdot)$ denotes a feature-extraction function.)}
	\vspace{-0.5em}
	\begin{tabular}{c|c|c}
		\hline
		Model & $\ell(\mathbf{X}, \mathbf{k}, \mathbf{v})$ & $\mathcal{R}(\mathbf{X})$ \\
		\hline \hline
		Linear attention~\cite{katharopoulos2020transformers} & $-\langle \mathbf{X} \mathbf{k}, \mathbf{v} \rangle$ & $-$ \\
		\hline
		Gated linear attention~\cite{yang2023gated} & $-\langle \mathbf{X} \mathbf{k}, \mathbf{v} \rangle$ & $\frac{1}{2} \| \mathrm{diag}(\sqrt{1 - \bm{\psi}}) \mathbf{X} \|_F^2$ \\
		\hline
		DeltaNet~\cite{schlag2021linear} & $\frac{1}{2} \| \mathbf{X} \mathbf{k} - \mathbf{v} \|^2$ & $-$  \\
		\hline
		Softmax attention w/o norm & $-\langle \mathbf{X} \phi(\mathbf{k}), \mathbf{v} \rangle$ & $-$ \\
		\hline
		Softmax attention w/ norm~\cite{vaswani2017attention} & 
		$- \langle \mathbf{X} \phi(\mathbf{k}), \mathbf{v} \rangle$ & $ \frac{1}{2} \| \mathbf{X} \|_F^2$  \\
		\hline
		Gated softmax attention~\cite{lin2025forgetting} & 
		$-\langle \mathbf{X} \phi(\mathbf{k}), \mathbf{v} \rangle$ & 
		$\frac{1}{2} \| \mathrm{diag}(\sqrt{1 - \bm{\psi}}) \mathbf{X} \|_F^2$  \\
		\hline
	\end{tabular}
	\label{tab:loss}
\end{table}

In order to account for information retrieval from each agent $m$,
the overall retrieval cost for agent $n$ includes all the costs $f_{m,t}(\mathbf{X}_{n,t})$, each weighted by a parameter $0  \le  w_{n , m} \le 1$, with $\sum_{m=1}^{N} w_{n , m} = 1$.
Accordingly, the weight $w_{n,m}$ dictates the relevance for agent $n$ of the data streamed at agent $m$.
Thus, the cumulative cost function at time $T$ for agent $n$ is given by the weighted sum
\begin{equation}\label{eq:1_loss}
	\mathcal{L}_{n}^T ( \mathbf{X}_{n}^T ) = \sum_{t=1}^T \sum_{m \in \mathcal{W}_n} w_{n , m} f_{m , t} ( \mathbf{X}_{n , t} ),
\end{equation}
where $\mathbf{X}_n^T = \{ \mathbf{X}_{n , t} \}_{t=1}^T$ is the set of AM parameters at agent $n$.

The cumulative cost \eqref{eq:1_loss} is determined by the logical relevance of local data among agents, which specifies whose data each agent needs to memorize. 
To summarize this \textit{logical connectivity} requirement, the logical weights $w_{n,m}$ are collected into the row-stochastic matrix $\mathbf{W}$ with $[\mathbf{W}]_{n , m} = w_{n , m}$, which is assumed to be known at all agents (see Fig.~\ref{fig:1}(b)). Accordingly, the set $\mathcal{W}_n = \{ m \in \mathcal{N} \mid w_{n,m} > 0 \}$ collects the agents whose data is logically relevant to agent~$n$.

Logical connectivity is distinct from \textit{physical connectivity}, which determines the available communication links between agents. 
In particular, agents communicate over an \textit{undirected graph} $\mathcal{G} = ( \mathcal{N} , \mathcal{E} )$ describing the physical connectivity among agents, where $\mathcal{N} = \{ 1 , 2 , \cdots , N\}$ is the set of agent indices, and $\mathcal{E} \subseteq  \mathcal{N} \times \mathcal{N}$ is a collection of inter-agent links $(m , n)$, indicating that agents $n$ and $m$ can directly exchange messages. We let $\mathcal{N}_n = \{ m \in \mathcal{N} : (m , n) \in \mathcal{E} \}$ denote the set of \textit{physical neighbors} of agent $n$.

Inter-agent communication on the physical network is generally essential to optimize the cumulative cost \eqref{eq:1_loss}. 
In fact, since the data $( \mathbf{k}_{m , t} , \mathbf{v}_{m , t} )$ are only available to agent $m$, the cost function $f_{m , t} ( \cdot )$ can only be evaluated by agent $m$. Therefore, in order to optimize the cost function \eqref{eq:1_loss}, as seen in Fig.~\ref{fig:1}(a), agent $n$ must generally communicate with other agents, unless agent $n$ only recalls local data, i.e., $w_{n,n} = 1$.

We emphasize that the logical relevance encoded by matrix $\mathbf{W}$ and the physical connectivity encoded by graph $\mathcal{G}$ are in general independent: an agent $m \in \mathcal{W}_n$ need not lie in $\mathcal{N}_n$, and vice versa. This mismatch is the central design challenge addressed by DDAM.

\vspace{-1em}
\subsection{Problem Definition}\label{Sec:2-B}
In this paper, we adopt a worst-case design approach, where optimality is evaluated with respect to the optimal solutions corresponding to the given sequences of keys and values observed by the agents.
Specifically, we investigate the following design objectives:
\begin{itemize}
	\item \textbf{Static regret:} The static regret is defined as the difference between the cumulative loss incurred by an online algorithm and the cumulative loss of the best fixed solutions $\{ \mathbf{U}_n^* \}_{n \in \mathcal{N}}$ in hindsight.
	The optimal solution for each agent $n$ is given by
	\begin{equation}\label{eq:best_solution}
		\mathbf{U}_n^* = \arg \min_{\mathbf{U} \in \mathcal{X}}  \mathcal{L}_{n}^T \left( \mathbf{U} \right) ,
	\end{equation}
	which is generally different across agents.
	Summing over the $N$ agents, the static regret is defined as
	\begin{equation}\label{eq:S_Reg}
		\begin{aligned}
			\mathrm{S}\text{-}\mathrm{Reg}(T) & = \sum_{n \in \mathcal{N}} \left( \mathcal{L}_{n}^T \left( \mathbf{X}_{n}^T \right)
			-  \mathcal{L}_{n}^T \left( \mathbf{U}_{n}^* \right) \right)  .
		\end{aligned}
	\end{equation}
	The static regret is relevant in stationary environments, for which there exists a single AM mechanism that can memorize well enough all data streamed within the time horizon $T$.

	\item \textbf{Dynamic regret:} The static regret is not an appropriate design criterion for changing environments, in which no single AM mechanism can retrieve data with sufficient accuracy over the entire time horizon of $T$ time steps. 
	In this case, the static regret in~\eqref{eq:S_Reg} may be negative, since there may exist online AM policies that outperform the best static solution in~\eqref{eq:best_solution}, even if the latter has access in hindsight to the entire data to be memorized.

	To address this limitation, we also study the dynamic regret \cite{zinkevich2003online,yang2016tracking,zhao2020dynamic,zhang2019adaptive}, in which the regret is evaluated against any sequence of changing AM mechanisms $\mathbf{U}_n^T = \{\mathbf{U}_{n,t}\}_{t=1}^T$, with $\mathbf{U}_{n,t}\in \mathcal{X}$ for all $t$ and $n\in\mathcal{N}$, as
	\begin{equation}\label{eq:D_Reg}
		\mathrm{D}\text{-}\mathrm{Reg}(T) = \sum_{n \in \mathcal{N}} \left( \mathcal{L}_{n}^T \left( \mathbf{X}_{n}^T \right)
		-  \mathcal{L}_{n}^T \left( \mathbf{U}_{n}^T \right) \right) .
	\end{equation}
	Note that static regret \eqref{eq:S_Reg} can be viewed as a special form of dynamic regret \eqref{eq:D_Reg} by setting the reference AM mechanism $\mathbf{U}_{n , t}$ as the fixed best solution in hindsight, i.e., $\mathbf{U}_{n , t} = \mathbf{U}_n^*$ for all time steps $t$.
\end{itemize}

\vspace{-1.51em}
\subsection{Technical Assumptions}
The analysis in this paper assumes the following standard conditions, which have also been adopted in, e.g., \cite{zinkevich2003online,yang2016tracking,zhao2020dynamic,zhang2019adaptive}.

\begin{assumption}\label{assump:0}
	The design domain $\mathcal{X} \subseteq \mathbb{R}^d$ is closed, convex, and it includes the origin, i.e., $\mathbf{0} \in \mathcal{X}$. 
	Furthermore, the cost function $f_{n , t} (\mathbf{X}_{n,t})$ is convex within the domain $\mathcal{X}$ at every time step $t$ for all $n \in \mathcal{N}$.
\end{assumption}
Note that all cost functions listed in Table~\ref{tab:loss} satisfy this assumption with $\mathcal{X}$ being the space of all matrices of suitable dimensions.

\begin{assumption}\label{assump:1}
	The diameter of the design domain $\mathcal{X}$ is upper bounded by $B < \infty$, i.e., $\|\mathbf{X} - \mathbf{Y}\| \leq B$ for all $\mathbf{X} , \mathbf{Y} \in \mathcal{X}$.
\end{assumption}
Based on Assumptions~\ref{assump:0} and \ref{assump:1}, it follows that the norm of the AM parameters $\mathbf{X}$ is bounded as $\|\mathbf{X}\|^2 \le B$.

\begin{assumption}\label{assump:2}
	For every time step $t$ and any $\mathbf{X} \in \mathcal{X}$, the gradient of the cost function $f_{n,t}(\cdot)$ is bounded in Euclidean norm by a constant $G_n$ with $0 <G_n < \infty$, i.e., $\left\| \nabla f_{n,t}(\mathbf{X}) \right\| \leq G_n $.
\end{assumption}

In addition to the above assumptions, we introduce the following definition of the path-length \cite{zinkevich2003online,yang2016tracking,zhao2020dynamic,zhang2019adaptive}, which measures the non-stationarity of the data streams.
\begin{definition}
	The path-length of the sequence $\mathbf{U}_n^T = \{\mathbf{U}_{n,t}\}_{t=1}^T$ of optimal solutions is defined as
	\begin{equation}\label{eq:path_length}
		\mathrm{PL}_n^T = \sum_{t=2}^{T} \| \mathbf{U}_{n, t-1} - \mathbf{U}_{n , t} \| .
	\end{equation}
\end{definition}
A larger path-length $\mathrm{PL}_n^T$ indicates that the data streams to be memorized at agent $n$ vary more significantly over time, requiring a higher temporal variability of the memory sequence at the agent.

\section{Conventional Protocols}\label{sec:3}
In this section, we review conventional solutions that can be leveraged to design DDAM protocols.
Specifically, we first study the ideal case in which every agent has access to full information, i.e., to all functions $\{ f_{n,t} (\cdot)\}_{n\in\mathcal{N}}$.
Then, we consider the special case of the DDAM problem introduced in Sec. \ref{sec:2}, in which agents assign the same weight to the data of all agents, i.e., $\mathbf{W} = \mathbbm{1} \mathbbm{1}^T / N$.
In this setting, one can impose without loss of optimality that all agents share identical AM parameters, so that distributed consensus protocols~\cite{Yan2013DisOCO} provide suitable solutions to the DDAM problem.

\vspace{-1em}
\subsection{Full-Information DDAM}\label{Sec:II-A}

For reference, we consider first an ideal full-information setting in which all agents have direct access to the data streams of all agents.  
In this case, each agent $n$ can evaluate the complete set of cost functions $\{ f_{n,t}(\cdot) \}_{n \in \mathcal{N}}$, and thus the cumulative cost \eqref{eq:1_loss}, at every time step $t$.
 
In this idealized setting, the problem of minimizing the static or dynamic regret can be addressed via OGD \cite{shalev2012online,orabona2019modern,zinkevich2003online}.
Specifically, OGD yields the update \cite{shalev2012online,orabona2019modern,zinkevich2003online}
\begin{equation}\label{eq:5}
	\mathbf{X}_{n,t+1} = \Pi_{\mathcal{X}} \Bigg[ \mathbf{X}_{n,t} - \eta_{n}  \sum_{m \in \mathcal{W}_n} w_{n , m}\nabla f_{m , t}( \mathbf{X}_{n,t} ) \Bigg],
\end{equation}
where $\Pi_{\mathcal{X}} [\cdot]$ denotes the projection onto the design set $\mathcal{X}$, and $\eta_{n}$ is the learning rate.

The regret properties of OGD follow directly from standard results \cite{orabona2019modern}.

\begin{theorem}\label{the:ogd}
	Under Assumptions~\ref{assump:0}–\ref{assump:2}, OGD attains the dynamic regret
	\begin{equation}\label{eq:ogd_reg}
		\mathrm{D}\text{-}\mathrm{Reg}(T) \le \sum_{n\in\mathcal{N}} \left( \frac{7B^2}{4\eta_{n}} + \frac{\eta_n T \bar{G}_n^2}{2} + \frac{B}{\eta_n} \mathrm{PL}_n^{T} \right) ,
	\end{equation}
	where we have written $\bar G_n \triangleq \sum_{m\in\mathcal{W}_n} w_{n,m} G_m$ and the
	path-length is defined in \eqref{eq:path_length}.
\end{theorem}

By setting an appropriate learning rate and selecting proper comparator sequences, the regret \eqref{eq:ogd_reg} can be further simplified, as stated in the following corollary.
\begin{corollary}\label{coro:ogd}
	By setting $\eta_n = B \sqrt{7} / \bar{G}_n \sqrt{2 T}$, the dynamic regret is upper bounded by
	\begin{equation}\label{eq:OGD_dym_reg}
		\begin{aligned}
			\mathrm{D}\text{-}\mathrm{Reg}(T) 
			& \le \sum_{n \in \mathcal{N}} \left( \sqrt{\frac{7}{2}} B \bar{G}_n \sqrt{T} + \sqrt{\frac{2}{7}} \bar{G}_n \mathrm{PL}_n^T \sqrt{T} \right) \\
			& = \mathcal{O} \left( \sum_{n \in \mathcal{N}} \left( 1 + \mathrm{PL}_n^T \right) \sqrt{T} \right) .
		\end{aligned}
	\end{equation}
	Furthermore, by letting $\mathbf{U}_{n , t} = \mathbf{U}_n^*$ in \eqref{eq:best_solution}, the path-length vanishes, i.e., $\mathrm{PL}_n^T = 0$, and the dynamic regret reduces to the static regret, yielding the upper bound
	\begin{equation}\label{eq:OGD_S_Reg}
		\begin{aligned}
			\mathrm{S}\text{-}\mathrm{Reg}(T) 
			& \le \sum_{n \in \mathcal{N}} \left( \sqrt{\frac{7}{2}} B \bar{G}_n \sqrt{T} \right) = \mathcal{O} \left( N \sqrt{T} \right) .
		\end{aligned}
	\end{equation}
\end{corollary}

Corollary~\ref{coro:ogd} demonstrates that OGD achieves a dynamic regret that scales with the path-length and thus with the degree of environmental variation.
In particular, if the data streams are stationary, i.e., the path-length $\mathrm{PL}_n^T = 0$, the dynamic regret reduces to the static regret \eqref{eq:OGD_S_Reg}, which is sublinear in the time horizon $T$.

The dynamic regret bound $\mathcal{O}(\sum_{n\in\mathcal{N}}(1 + \mathrm{PL}_n^T)\sqrt{T})$ in \eqref{eq:OGD_dym_reg} matches the standard rate achieved by OGD \cite{zinkevich2003online}. 
Tighter bounds of order $\mathcal{O}( \sqrt{\mathrm{PL}_n^{\raisebox{-0.2ex}{$\scriptstyle T$}} T} )$ are known to be achievable for convex losses, but require prior knowledge of the path-length $\mathrm{PL}_n^T$ \cite{zhang2018adaptive}.
Since the path-length $\mathrm{PL}_n^T$ is generally unknown in advance, we focus on the standard bound in \eqref{eq:OGD_dym_reg}, which holds without any prior knowledge of the true variability of the data streams.

Since OGD assumes that each agent has access to full information, it is a centralized scheme that cannot be implemented in the distributed setting of interest. 
We thus include it as an \emph{oracle} baseline, and the regret bounds in Corollary \ref{coro:ogd} serve as an idealized full-information benchmark for decentralized protocols.

\subsection{Consensus DDAM}\label{C-DOGD}\label{Sec:II-B}
In this subsection, we consider a special case of the DDAM problem introduced in the previous section in which all the logical weights in the cost function \eqref{eq:1_loss} are equal, i.e., the weight matrix is set to $\mathbf{W}=\mathbbm{1}\mathbbm{1}^T/N$.
In this setting, the cost functions in~\eqref{eq:1_loss} become identical for all agents $n \in \mathcal{N}$.
Therefore, there exists a shared optimal AM mechanism $\mathbf{U}^*$ for all agents, i.e., $\mathbf{U}_n^* = \mathbf{U}^*$.

Under the consensus protocol, all agents asymptotically agree on a common solution~$\mathbf{X}$, i.e., $\mathbf{X}_{n,t} \to \mathbf{X}$ as $t \to \infty$.
Since this comes with no loss of generality in this setting, we focus here on \textit{consensus distributed OGD} (C-DOGD) \cite{Yan2013DisOCO} to address the DDAM problem in this special case.
C-DOGD is specified by an $N \times N$ doubly stochastic matrix $\mathbf{A}$ with entries $[\mathbf{A}]_{n , m} = a_{n,m}$ for all $n\in\mathcal{N}$ and $m\in\mathcal{N}$. 
In particular, C-DOGD performs local gradient steps followed by consensus averaging as
\begin{equation}\label{eq:C_DOGD}
	\mathbf{X}_{n,t+1} = \Pi_{\mathcal{X}} \left[\sum_{m \in \mathcal{N}_n} a_{n , m} \mathbf{X}_{m,t} - \eta_{n} \nabla f_{n , t} (\mathbf{X}_{n , t}) \right],
\end{equation}
where $\eta_{n}$ is the learning rate.

To implement the update \eqref{eq:C_DOGD}, each agent $n$ needs to communicate with its neighbors in the set $\mathcal{N}_n$ to exchange the local AM parameters.
As a result, all edges in the graph $\mathcal{G}$ are activated, each sustaining a communication load of 
\begin{equation}\label{eq:12_c_dogd_capacity}
	C_{\max} = 2
\end{equation}
AM parameters per iteration.

The static regret properties of C-DOGD follow directly from \cite{Yan2013DisOCO}.
\begin{theorem}
	In the special case $\mathbf{W} = \mathbbm{1} \mathbbm{1}^T / N$, under Assumptions~\ref{assump:0}–\ref{assump:2}, with $\eta_{n} = {1} / {2 \sqrt{T}}$, the C-DOGD protocol obtains the static regret \cite[Theorem 1]{Yan2013DisOCO}
	\begin{equation}\label{eq:c_dogd}
		\mathrm{S}\text{-}\mathrm{Reg}(T) \le N \left(B + \frac{5 - \alpha}{1 - \alpha} G_{\max}^2\right)\sqrt{T} 
		= \mathcal{O}\left( N \sqrt{T} \right) ,
	\end{equation}
	where we have defined $G_{\max} = \max_{n \in \mathcal{N}} G_n$ and $\alpha$ is the spectral gap of the doubly stochastic matrix $\mathbf{A}$.
\end{theorem}

Although C-DOGD circumvents the need for global loss information, its sublinear static regret guarantee in \eqref{eq:c_dogd} holds only in the special case where all agents aim to memorize the same information.
Furthermore, to the best of our knowledge, the dynamic regret performance of C-DOGD has yet to be investigated.

\section{Distributed Dynamic Associative Memory via Tree-based Online Gradient Descent}\label{sec:4}
In this section, we study the DDAM problem introduced in Sec.~\ref{sec:2} in its full generality by proposing a new protocol, \textit{DDAM via tree-based OGD} (DDAM-TOGD), which will be demonstrated to exhibit guaranteed regret properties.

\vspace{-1em}
\subsection{DDAM-TOGD}

In the DDAM problem introduced in Sec. \ref{sec:2}, each agent $n$ aims at memorizing data from a subset $\mathcal{W}_n$ of agents.  
The consensus-based distributed protocol introduced in Sec.~\ref{Sec:II-B} relies on neighbor-only communication followed by local averaging, which is appropriate when all agents share a common memory. 
However, DDAM requires each agent $n$ to preserve selective information from possibly non-neighboring agents in $\mathcal{W}_n$.
To this end, messages must be transmitted along dedicated routes, avoiding the intermediate mixing required by protocols based on neighbor-only communication.

DDAM-TOGD addresses this problem by enabling inter-agent communication over routing trees connecting each agent $n$ to all relevant agents in $\mathcal{W}_n \backslash \{n\}$ on the physical connectivity graph $\mathcal{G}$.
Trees minimize the number of activated edges and avoid redundant transmissions while still providing a unique path between $n$ and each target for broadcast transmissions \cite{west2001introduction,newman2018networks}.
Based on the delayed information received from the agents in subset $\mathcal{W}_n \backslash \{n\}$, each agent $n$ applies a form of OGD on the cumulative cost in \eqref{eq:1_loss}.

To elaborate, given the physical communication graph $\mathcal{G}$ and the logical weight matrix $\mathbf{W}$, DDAM-TOGD first constructs $N$ routing trees $\{ \mathcal{T}_n \}_{n \in \mathcal{N}}$.
Each tree $\mathcal{T}_n$ is rooted at agent $n$ and contains paths, i.e., sequences of contiguous edges in graph $\mathcal{G}$, from agent $n$ to all agents in subset $\mathcal{W}_n$.
The construction of the trees is arbitrary, but we will introduce an optimized design based on the regret analysis in Sec.~\ref{Sec-III-C}.

At each time step $t$, agent $n$ sends its current iterate $\mathbf{X}_{n , t}$ along tree $\mathcal{T}_n$ toward all agents $m \in \mathcal{W}_n \backslash \{n\}$.
This information requires $\tilde{\tau}_{n , m}$ time steps to arrive at agent $m$, where $\tilde{\tau}_{n , m}$ is the number of edges on the path from agent $n$ to agent $m$.
Upon receiving the query at time $t + \tilde{\tau}_{n , m}$, agent $m$ evaluates the gradient $\nabla f_{m , t} ( \mathbf{X}_{n , t} )$ of the cost corresponding to the data $( \mathbf{k}_{m,t} , \mathbf{v}_{m,t} )$, and sends it back along the reverse path.
We denote as $\tau_{n , m} = 2 \tilde{\tau}_{n , m}$ the total communication delay incurred in the round-trip exchange between agent $n$ and agent $m$.

Accordingly, agent $n$ receives the gradient $\nabla f_{m , t} (\mathbf{X}_{n , t})$ only at time $t + \tau_{n , m}$.
Thus, at any time $t$ it does not have access to the current gradients from other agents and can only rely on delayed gradients $\nabla f_{m , t - \tau_{n , m}} ( \mathbf{X}_{n , t - \tau_{n , m}} )$ from each target agent $m \in \mathcal{W}_n \backslash \{n\}$. 
Using the most recent gradient information available from the agents in the subset $\mathcal{W}_n \backslash \{n\}$, at time step $t+1$, each agent applies the update
\begin{equation}\label{eq:6}
	\mathbf{X}_{n,t+1} = \Pi_{\mathcal{X}} \Bigg[ \mathbf{X}_{n,t}   -  \eta_{n}  \sum_{m \in \mathcal{W}_n}    w_{n , m}  \mathbf{G}_{m , t - \tau_{n,m}}^n \mathbbm{1}_{\{ t > \tau_{n , m} \}} \Bigg] ,
\end{equation}
where $\mathbf{G}_{m , t - \tau_{n,m}}^n = \nabla f_{m , t-\tau_{n,m}}( \mathbf{X}_{n,t-\tau_{n,m}} )$, and $\eta_{n}$ is the learning rate.
In \eqref{eq:6}, the gradient corresponding to agent $m$ is multiplied by the weight $w_{n,m}$, 
reflecting the structure of the cumulative cost function \eqref{eq:1_loss}.
The overall DDAM-TOGD is summarized in Algorithm \ref{alg:1}.

For each agent $n$ and each edge $e \in \mathcal{E}$, let $C_{n, m, e} \in \{0, 1\}$ be a binary variable indicating whether the routing tree $\mathcal{T}_n$ includes edge $e$ for delivering a message from the root node $n$ to a target agent $m$. Specifically, $C_{n, m, e} = 1$ if edge $e$ is used in the path from $n$ to $m$, and $C_{n, m, e} = 0$ otherwise.
The overall maximum per-link capacity required by DDAM-TOGD for a given choice $\mathcal{T} = \{ \mathcal{T}_n \}$ of trees is thus given by
\begin{equation}\label{eq:capacity_max}
	C_{\max}^{\mathcal{T}} = 2 \max_{e \in \mathcal{E}} \sum_{n \in \mathcal{N}} \sum_{m \in \mathcal{W}_n} C_{n , m , e} .
\end{equation}
Note that this is generally larger than the per-link capacity \eqref{eq:12_c_dogd_capacity} required by C-DOGD.

\begin{algorithm}[t]
	\renewcommand{\algorithmicrequire}{\textbf{Input:}}
	\renewcommand{\algorithmicensure}{For each agent $n$, apply the following routine:}
	\caption{DDAM-TOGD (at agent $n$)}
	\begin{algorithmic}[1] 
		%		\ENSURE 
		\REQUIRE Learning rate sequences $\{ \eta_{n} \}_{n \in \mathcal{N}}$, logical weight $\{ w_{n , m}\}_{m \in \mathcal{W}_n}$, and tree $\mathcal{T}_n$
		%		\STATE For each agent $n$, apply the following routine:\;
		\STATE \textbf{Initialize} AM parameters $\{\mathbf{X}_{n,1}\}_{n\in\mathcal{N}}$
		\FOR{$t = 1 , 2 , \dots$}
		\STATE \textcolor{blue}{\# \textit{Receive memory parameters and send gradients}}
		\STATE Receive $\mathbf{X}_{m , t - \tilde{\tau}_{m , n}}$ from agent $m$ \;
		\STATE Send $\nabla f_{n , t - \tilde{\tau}_{m , n}}(\mathbf{X}_{m , t - \tilde{\tau}_{m , n}})$ to agent $m$ \;
		\STATE \textcolor{blue}{\# \textit{Send memory parameter and receive gradients}}
		\STATE Broadcast $\mathbf{X}_{n , t}$ to all agents $m \in \mathcal{W}_n \backslash \{n\}$ \;
		\STATE Receive {$\nabla f_{m , t  - \tau_{n , m}} ( \mathbf{X}_{n, t - \tau_{n , m}} )$} from all agents {$m \in \mathcal{W}_n  \backslash  \{ n\}$}
		\STATE \textcolor{blue}{\# \textit{Online Gradient Descent}}
		\STATE Update local memory parameter using \eqref{eq:6}
		\ENDFOR
	\end{algorithmic}
	\label{alg:1}
\end{algorithm}

\vspace{-1em}
\subsection{Regret Analysis}
The regret of the proposed DDAM-TOGD algorithm is characterized in the following theorem.
\begin{theorem}\label{Theo:DAM-TOGD}
	Under Assumptions~\ref{assump:0}–\ref{assump:2}, the dynamic regret of the proposed DDAM-TOGD protocol satisfies the upper bound
	\begin{equation}
		\begin{aligned}
			\mathrm{D\text{-}Reg}(T) \le & \sum_{n\in\mathcal{N}} \Bigg( Q_n \eta_{n} \left( T + \Delta \tau_n \right) 
			+ J_n \eta_{n} 
			+ \frac{7 B^2}{4 \eta_n} \\
			&  \qquad + \left( \frac{B}{\eta_{n}} + H_n \right) \mathrm{PL}_n^{T+\tau_{n,\max}}  
			+ C_n \Bigg) ,
		\end{aligned}
	\end{equation}
	where 
	\begin{align}
		Q_n = & \frac{1}{2} K_n \left( \sum_{m \in \mathcal{W}_n}  G_m \right) + |\mathcal{W}_n| K_n^2 \tau_{n,\mathrm{sum}} , \\
		J_n = & |\mathcal{W}_n|^2 K_n^2  \tau_{n, \max}^2 , \\
		H_n = & K_n \tau_{n,\mathrm{sum}} , \\
		C_n = & K_n \Delta \tau_n |\mathcal{W}_n| B ,
	\end{align}
	with
	\begin{align}
		K_n & = \max_{ m } w_{n , m} G_m , \label{eq:20}\\
		\tau_{n,\min} & = \min_m \tau_{n , m} , \label{eq:21}\\
		\tau_{n,\max} & = \max_m \tau_{n , m} , \label{eq:22}\\
		\Delta \tau_n & = \tau_{n , \max} - \tau_{n , \min} , \label{eq:23}\\
		\tau_{n,\mathrm{sum}} & = \sum_{m \in \mathcal{W}_n} \tau_{n , m} . \label{eq:nn24}
	\end{align}
\end{theorem}
The proof of Theorem~\ref{Theo:DAM-TOGD} is discussed in Sec.~\ref{sec:5}.
Based on Theorem~\ref{Theo:DAM-TOGD}, by specifying the learning rate, the static and dynamic regrets can be specialized as stated in the following corollary.
\begin{corollary}\label{Coro:D_Reg_Conv}	
	By setting $\eta_n \!=\! \sqrt{ 7 B^2 / 4 \left( Q_n(T+\Delta \tau_n) + J_n\right) }$, the dynamic regret of proposed DDAM-TOGD is upper bounded by
	\begin{align}
		& \mathrm{D}\text{-}\mathrm{Reg}(T) \notag \\
		& \le \sum_{n \in \mathcal{N}} \Bigg( \sqrt{7} B \sqrt{Q_n(T+\Delta\tau_n)+J_n}  \notag \\
		& \quad + \left( H_n + \frac{2}{\sqrt{7}} \sqrt{Q_n(T+\Delta\tau_n)+J_n}  \right) \mathrm{PL}_n^{T+\tau_{n,\max}} + C_n
		\Bigg) \notag \\
		& = \mathcal{O} \!\! \left( \! \sum_{n \in \mathcal{N}} \! \! \left(  \!\left(  1 \!+\! \mathrm{PL}_n^{T+\tau_{n,\max}} \right) \! \sqrt{\left(1 + \tau_{n,\mathrm{sum}} \right) (T + \Delta \tau_n)} \right) \right) . \label{eq:DAM-TOGD-Regret-1}
	\end{align}
	Furthermore, by letting $\mathbf{U}_{n , t} = \mathbf{U}_n^*$ in \eqref{eq:best_solution}, the path-length vanishes, i.e., $\mathrm{PL}_n^T = 0$, and the dynamic regret reduces to the static regret, yielding the upper bound
	\begin{equation}\label{eq:DAM-TOGD-Regret-2}
		\begin{aligned}
			\mathrm{S}\text{-}\mathrm{Reg}(T) 
			& \le \sum_{n \in \mathcal{N}} \sqrt{7} B \sqrt{Q_n(T+\Delta\tau_n)+J_n} + C_n \\
			& = \mathcal{O} \left( \sum_{n \in \mathcal{N}} \sqrt{\left(1 + \tau_{n,\mathrm{sum}} \right) (T + \Delta \tau_n)} \right) .
		\end{aligned}
	\end{equation}
\end{corollary}

Compared with the idealized OGD bound in \eqref{eq:OGD_dym_reg}, where the path-length is scaled by $\sqrt{T}$, the dynamic regret bound of DDAM-TOGD contains the product between path-length and the term  $\sqrt{(1+\tau_{n,\mathrm{sum}})(T+\Delta\tau_n)}$, making the regret explicitly dependent on the communication delay. 
Here, $\tau_{n,\mathrm{sum}}$ in \eqref{eq:nn24} is the cumulative delay from agent $n$ to all agents of interest in $\mathcal{W}_n$, and $\Delta\tau_n$ in \eqref{eq:23} is the difference between the maximum and minimum delays from agent $n$. 
This delay dependence reflects the fact that information exchange in the distributed network slows down the learning progress of the AM process.

This result reveals a fundamental trade-off between the non-stationarity of the data streams and the communication delay. 
Specifically, the path-length $\mathrm{PL}_n^{T+\tau_{n,\max}}$ grows with the variability of the optimal memory sequence, while the delay terms $\tau_{n,\mathrm{sum}}$ and $\Delta\tau_n$ scale with the size of the routing trees. 
For a fixed target regret level, a more pronounced non-stationarity can therefore be supported only on a smaller network, characterized by a smaller delay. 
This fundamental limitation motivates the communication-efficient tree design in the next subsection, and is further validated empirically in Sec.~\ref{sec:6}.

That said when the time horizon $T$ is much larger than the sum-delay $\tau_{n,\mathrm{sum}}$, the coefficient $\sqrt{(1+\tau_{n,\mathrm{sum}})(T+\Delta\tau_n)}$ remains sublinear in $T$, and thus DDAM-TOGD recovers the same sublinear time dependence as the idealized OGD. 
The same conclusion holds for the static regret \eqref{eq:DAM-TOGD-Regret-2}.

\vspace{-1em}
\subsection{Design of Communication Trees}\label{Sec-III-C}

The proposed DDAM-TOGD protocol applies to any choice of the routing trees $\mathcal{T}_n$ connecting each agent $n$ to all the agents $m \in \mathcal{W}_n$, whose data must be memorized by agent $n$. 
However, the regret bounds in \eqref{eq:DAM-TOGD-Regret-1} and \eqref{eq:DAM-TOGD-Regret-2} provide useful insights into the design of the routing trees $\{\mathcal{T}_n\}_{n \in \mathcal{N}}$. 
In particular, the upper bound highlights the influence of the term $(1 + \tau_{n , \mathrm{sum}}) (T + \Delta \tau_n)$ on each agent's performance.
As discussed in the previous subsection, when the memorization horizon $T$ is large enough, the dominant component of the regret bound is the cumulative delay term $\tau_{n , \mathrm{sum}}$.
Based on these insights, we propose to design the trees $\mathcal{T}_n$ with the aim of minimizing the sum-delay $\tau_{n , \mathrm{sum}}$.
The proposed combinatorial optimization procedure is detailed in Appendix~\ref{App:Tree_Optim}, and we refer to the resulting scheme as $\mathrm{DDAM}\text{-}\mathrm{TOGD}^\star$.

\section{Theoretical Analysis}\label{sec:5}
In this section, we outline the proof of Theorem~\ref{Theo:DAM-TOGD}, with additional details provided in the Appendix.
The proof proceeds by first decomposing the dynamic regret \eqref{eq:D_Reg} to isolate the impact of communication delays and then bounding each constituent term separately.

\vspace{-1em}
\subsection{Regret Decomposition}
The network dynamic regret \eqref{eq:D_Reg} can be expressed as the sum of the local dynamic regrets of all agents as
\begin{equation}
	\begin{aligned}
		\mathrm{D\text{-}Reg}(T) & = \sum_{n \in \mathcal{N}} \Big( \underbrace{ \mathcal{L}_{n}^T \left( \mathbf{X}_{n}^T \right)
			-  \mathcal{L}_{n}^T \left( \mathbf{U}_{n}^T \right)}_{\mathrm{D\text{-}Reg}_n(T)} \Big) \\
		& = \sum_{n \in \mathcal{N}} \mathrm{D\text{-}Reg}_n(T) .
	\end{aligned}
\end{equation}
We next derive a bound for the dynamic regret of each agent $n$, i.e., $\mathrm{D\text{-}Reg}_n(T)$. 
Summing these bounds over all agents yields an overall network-level guarantee. 
To start, the local dynamic regret of agent $n$ can be upper bounded in a convenient way that isolates the contributions of communication delays
\begin{align}
	& \mathrm{D\text{-}Reg}_n(T) = \mathcal{L}_{n}^T \left( \mathbf{X}_{n}^T \right)
	-  \mathcal{L}_{n}^T \left( \mathbf{U}_{n}^T \right) \notag \\
	& \mathop \le \limits^{\text{(a)}} \sum_{t=1}^{T} \sum_{m \in \mathcal{W}_n} w_{n , m}   \left\langle\mathbf{G}_{m , t}^n  , \mathbf{X}_{n , t} - \mathbf{U}_{n,t} \right\rangle \notag \\
	& \mathop = \limits^{\text{(b)}} \!\!\! \sum_{m \in \mathcal{W}_n} \sum_{t=\tau_{n ,m}+1}^{T+\tau_{n, m}} \!\!\! w_{n , m} \left\langle\mathbf{G}_{m , t-\tau_{n, m}}^n  , \mathbf{X}_{n , t-\tau_{n, m}} - \mathbf{U}_{n,t-\tau_{n, m}} \right\rangle  \notag \\ 
	& = \!\!\! \underbrace{ \sum_{t=\tau_{n ,\min}+1}^{T+\tau_{n, \max}} \sum_{m \in \mathcal{W}_n}  \!\!\! w_{n , m} \left\langle\mathbf{G}_{m , t-\tau_{n, m}}^n \mathbbm{1}_{\{t > \tau_{n,m}\}} , \mathbf{X}_{n , t} - \mathbf{U}_{n,t} \right\rangle }_{ \mathtt{Reg^{\star}} }  \notag \\ 
	& \quad + \underbrace{ \sum_{m \in \mathcal{W}_n} \sum_{t=\tau_{n ,m}+1}^{T+\tau_{n, m}} w_{n , m} \left\langle\mathbf{G}_{m , t-\tau_{n, m}}^n  , \mathbf{X}_{n , t-\tau_{n, m}} - \mathbf{X}_{n,t} \right\rangle }_{ \mathtt{Drift}_{\mathbf{X}} } \notag\\ 
	& \quad + \underbrace{ \sum_{m \in \mathcal{W}_n} \sum_{t=\tau_{n ,m}+1}^{T+\tau_{n, m}} w_{n , m} \left\langle\mathbf{G}_{m , t-\tau_{n, m}}^n  , \mathbf{U}_{n , t} - \mathbf{U}_{n,t-\tau_{n, m}} \right\rangle }_{ \mathtt{Drift}_{\mathbf{U}} } \notag \\ 
	& \quad +  \underbrace{ \sum_{m \in \mathcal{W}_n} \sum_{t=T+\tau_{n ,m}+1}^{T+\tau_{n, \max}} w_{n , m} \left\langle\mathbf{G}_{m , t-\tau_{n, m}}^n  , \mathbf{U}_{n,t} - \mathbf{X}_{n , t} \right\rangle }_{ \mathtt{Tail} } , \label{eq:reg_dec_1}
\end{align}
where (a) holds by the convexity of the local loss functions \cite[Lemma 2.5]{shalev2012online}, and (b) holds by changing the order of summation and adjusting the start and end indices.
In the decomposition \eqref{eq:reg_dec_1}, the term $\mathtt{Reg}^\star$ retains the characteristic form of conventional regret analyses \cite{orabona2019modern}, while the remaining terms, namely $\mathtt{Drift}_{\mathbf{X}}$, $\mathtt{Drift}_{\mathbf{U}}$, and $\mathtt{Tail}$, are consequences of the presence of communication delays.
Specifically, the terms $\mathtt{Drift}_{\mathbf{X}}$ and $\mathtt{Drift}_{\mathbf{U}}$ quantify the drift caused by the mismatch between stale and current iterates in the decision and comparator sequences, respectively; while the term $\mathtt{Tail}$ accounts for gradient information that arrives only after the horizon $T$ and hence cannot be incorporated into the updates.

\vspace{-1em}
\subsection{Bounding Individual Terms}
We now analyze each term separately.
Combining the bounds below immediately yields Theorem~\ref{Theo:DAM-TOGD}.

\subsubsection{Bounding $\mathtt{Reg}^\star$}
Following \cite[Lemma 2.12]{orabona2019modern}, we first establish the following key lemma.
\begin{lemma}\label{lem:single_step}
	Under Assumption \ref{assump:2}, the DDAM-TOGD update rule in~\eqref{eq:6} satisfies the following inequality
	\begin{equation}\label{eq:lemm_23}
		\begin{aligned}
			& \sum_{m \in \mathcal{W}_n} \!\!\! \left\langle w_{n , m} \mathbf{G}_{m , t-\tau_{n , m}}^n \mathbbm{1}_{\{t > \tau_{n , m}\}} , \mathbf{X}_{n , t} - \mathbf{U}_{n,t} \right\rangle  \\
			& \le \frac{\left\| \mathbf{X}_{n , t} - \mathbf{U}_{n,t} \right\|^2 - \left\| \mathbf{X}_{n , t+1} - \mathbf{U}_{n,t} \right\|^2}{2 \eta_{n}} \\
			& \quad + K_n \left( \sum_{m \in \mathcal{W}_n}  G_m \right)  \frac{\eta_{n}}{2},
		\end{aligned}
	\end{equation}
	where $K_n$ is defined in \eqref{eq:20}.
\end{lemma}
\begin{IEEEproof}
	See Appendix \ref{App:lem:single_step}.
\end{IEEEproof}

Summing \eqref{eq:lemm_23} over index $t$ from $\tau_{\min}$ to $T + \tau_{\max}$ gives the upper bound
\begin{align}
	\mathtt{Reg}^\star = & \!\!\!\!\!\! \sum_{t=\tau_{n ,\min}+1}^{T+\tau_{n, \max}} \sum_{m \in \mathcal{W}_n} \!\!\!\! w_{n , m} \left\langle\mathbf{G}_{m , t-\tau_{n, m}}^n \mathbbm{1}_{\{t > \tau_{n,m}\}} , \mathbf{X}_{n , t} \!\! - \!\! \mathbf{U}_{n,t} \right\rangle \notag \\
	\le   & \frac{1}{2} K_n \left( \sum_{m \in \mathcal{W}_n}  G_m \right) \eta_n \left(T + \Delta \tau_n\right) \notag \\
	& + \sum_{t=\tau_{n ,\min}+1}^{T+\tau_{n, \max}} \frac{\left\| \mathbf{X}_{n , t} - \mathbf{U}_{n,t} \right\|^2 - \left\| \mathbf{X}_{n , t+1} - \mathbf{U}_{n,t} \right\|^2}{2 \eta_{n}} . \label{eq:reg_star_1}
\end{align}
The telescoping term in \eqref{eq:reg_star_1} can be bounded as follows.
\begin{lemma}\label{lem:telescoping}
	Under Assumption \ref{assump:1}, the telescoping term in~\eqref{eq:reg_star_1} can be bounded as
	\begin{equation}\label{eq:reg_star_2}
		\begin{aligned}
			& \sum_{t=\tau_{n ,\min}+1}^{T+\tau_{n, \max}} \frac{\left\| \mathbf{X}_{n , t} - \mathbf{U}_{n,t} \right\|^2 - \left\| \mathbf{X}_{n , t+1} - \mathbf{U}_{n,t} \right\|^2}{2 \eta_{n}} \\
			& \le \frac{B}{\eta_n} \mathrm{PL}_n^{T+\tau_{n,\max}} + \frac{7B^2}{4\eta_n} .
		\end{aligned}
	\end{equation}
\end{lemma}
\begin{IEEEproof}
	See Appendix \ref{App:lem:telescoping}.
\end{IEEEproof}

Substituting the telescoping bound in Lemma~\ref{lem:telescoping} into~\eqref{eq:reg_star_1} directly yields the following bound for $\mathtt{Reg}^{\star}$.
\begin{proposition}[Bound on $\mathtt{Reg}^{\star}$]
	Under Assumptions~\ref{assump:1} and~\ref{assump:2}, the term $\mathtt{Reg}^{\star}$ satisfies the upper bound
	\begin{equation}\label{eq:n_t_1}
		\begin{aligned}
			\mathtt{Reg}^{\star} \le & \frac{1}{2} K_n \left( \sum_{m \in \mathcal{W}_n}  G_m \right) \eta_n \left(T + \Delta \tau_n\right)    \\
			& + \frac{B}{\eta_n} \mathrm{PL}_n^{T+\tau_{n,\max}} + \frac{7B^2}{4\eta_n}.
		\end{aligned}
	\end{equation}
\end{proposition}
This concludes the proof of the bound on the term $\mathtt{Reg}^{\star}$. 
We next bound the term $\mathtt{Drift}_{\mathbf{X}}$.

\subsubsection{Bounding $\mathtt{Drift}_{\mathbf{X}}$}

By applying the Cauchy-Schwarz inequality together with Assumption \ref{assump:2}, $\mathtt{Drift}_{\mathbf{X}}$ is bounded as
\begin{align}
	\mathtt{Drift}_{\mathbf{X}} & \! = \!\! \sum_{m \in \mathcal{W}_n} \sum_{t=\tau_{n ,m}+1}^{T+\tau_{n, m}} \!\!\!\! w_{n , m} \left\langle\mathbf{G}_{m , t-\tau_{n, m}}^n  , \mathbf{X}_{n , t-\tau_{n, m}} \!\! - \!\! \mathbf{X}_{n,t} \right\rangle \notag \\
	& \le \sum_{m \in \mathcal{W}_n} \sum_{t=\tau_{n ,m}+1}^{T+\tau_{n, m}} w_{n , m} G_m \left\| \mathbf{X}_{n,t} - \mathbf{X}_{n , t-\tau_{n, m}} \right\| \notag \\
	& \le \sum_{m \in \mathcal{W}_n} \sum_{t=\tau_{n ,m}+1}^{T+\tau_{n, m}} K_n \left\| \mathbf{X}_{n,t} - \mathbf{X}_{n , t-\tau_{n, m}} \right\| . \label{eq:reg_drift_1}
\end{align}

Thus, the bound of $\mathtt{Drift}_{\mathbf{X}}$ depends on the term $\left\| \mathbf{X}_{n,t} - \mathbf{X}_{n , t-\tau_{n, m}} \right\|$, which measures the difference between the variables with and without delay.
The following lemma bounds this difference.
\begin{lemma}\label{lem:bound_x_difference}
	Under Assumption~\ref{assump:2}, for any $t > \tau_{n,m}$, it holds that
	\begin{equation}\label{eq:reg_drift_2_Main}
		\begin{aligned}
			& \left\| \mathbf{X}_{n , t} - \mathbf{X}_{n , t -\tau_{n , m}} \right\| 
			= \left\| \sum_{j=1}^{\tau_{n , m}} ( \mathbf{X}_{n , t-j+1} - \mathbf{X}_{n , t-j} ) \right\| \\
			& \le  \eta_{n} K_n \sum_{s \in \mathcal{W}_n} \min \{ \tau_{n , m} , t-\tau_{n , s}-1  \} .
		\end{aligned}
	\end{equation}
\end{lemma}
\begin{IEEEproof}
	See Appendix \ref{App:eq:reg_drift_2}. 
\end{IEEEproof}

Substituting Lemma~\ref{lem:bound_x_difference} into \eqref{eq:reg_drift_1} and simplifying yield the following proposition.
\begin{proposition}[Bound on $\mathtt{Drift}_{\mathbf{X}}$]
	Under Assumptions~\ref{assump:1} and~\ref{assump:2}, the term $\mathtt{Drift}_{\mathbf{X}}$ satisfies the upper bound
	\begin{equation}\label{eq:reg_drift_3}
		\begin{aligned}
			\mathtt{Drift}_{\mathbf{X}} \le & |\mathcal{W}_n|^2 K_n^2  \tau_{n, \max}^2 \eta_{n} \\
			& + |\mathcal{W}_n| K_n^2 \tau_{n,\mathrm{sum}} \eta_{n} \left( T + \Delta \tau_n \right) .
		\end{aligned}
	\end{equation}
\end{proposition}
\begin{IEEEproof}
	See Appendix~\ref{App:eq:reg_drift_2}.
\end{IEEEproof}

\subsubsection{Bounding $\mathtt{Drift}_{\mathbf{U}}$}
By applying the Cauchy-Schwarz inequality together with Assumption \ref{assump:2}, we obtain
\begin{align}
		\mathtt{Drift}_{\mathbf{U}} & = \!\!\! \sum_{m \in \mathcal{W}_n} \sum_{t=\tau_{n ,m}+1}^{T+\tau_{n, m}} \!\!\! w_{n , m} \left\langle\mathbf{G}_{m , t-\tau_{n, m}}^n  , \mathbf{U}_{n , t} \!-\! \mathbf{U}_{n,t-\tau_{n, m}} \right\rangle \notag \\
		& \le \sum_{m \in \mathcal{W}_n} \sum_{t=\tau_{n ,m}+1}^{T+\tau_{n, m}} w_{n , m} G_m \left\| \mathbf{U}_{n , t} - \mathbf{U}_{n,t-\tau_{n, m}} \right\| \notag \\
		& \le K_n \sum_{m \in \mathcal{W}_n} \sum_{t=\tau_{n ,m}+1}^{T+\tau_{n, m}} \left\| \mathbf{U}_{n , t} - \mathbf{U}_{n,t-\tau_{n, m}} \right\| . \label{eq:reg_drift2_1}
\end{align}
The expression \eqref{eq:reg_drift2_1} reveals that the term $\mathtt{Drift}_{\mathbf{U}}$ arises from the discrepancy between the comparator sequences with and without delay.
\begin{lemma}\label{lemm:drift_2}
	The drift of the comparator caused by delay can be bounded as
	\begin{equation}
		\sum_{t=\tau_{n ,m}+1}^{T+\tau_{n, m}} \left\| \mathbf{U}_{n , t} - \mathbf{U}_{n,t-\tau_{n, m}} \right\| \le \tau_{n,m} \mathrm{PL}_n^{T + \tau_{n , m}} .
	\end{equation}
\end{lemma}
\begin{IEEEproof}
	See Appendix \ref{App:drift_2}. 
\end{IEEEproof}

Combining~\eqref{eq:reg_drift2_1} with Lemma~\ref{lemm:drift_2} 
yields the following bound.
\begin{proposition}[Bound on $\mathtt{Drift}_{\mathbf{U}}$]\label{prop:drift_U}
	Under Assumption~\ref{assump:2}, the term $\mathtt{Drift}_{\mathbf{U}}$ 
	satisfies the upper bound
	\begin{equation}\label{eq:reg_drift2_final}
		\mathtt{Drift}_{\mathbf{U}} 
		\le K_n \tau_{n,\mathrm{sum}}\, \mathrm{PL}_n^{T+\tau_{n,\max}}.
	\end{equation}
\end{proposition}
\begin{IEEEproof}
	Substituting Lemma~\ref{lemm:drift_2} into~\eqref{eq:reg_drift2_1}, 
	\begin{equation}
		\begin{aligned}
			\mathtt{Drift}_{\mathbf{U}} & \le K_n \sum_{m \in \mathcal{W}_n} \tau_{n,m} \mathrm{PL}_n^{T + \tau_{n,m}} \\
			& \mathop \le \limits^{(\text{a})} K_n \tau_{n,\mathrm{sum}} \mathrm{PL}_n^{T + \tau_{n,\max}} ,
		\end{aligned}
	\end{equation}
	where (a) follows from the inequality $\mathrm{PL}_n^{T + \tau_{n,m}} \le \mathrm{PL}_n^{T + \tau_{n,\max}}$ for all $m \in \mathcal{W}_n$.
\end{IEEEproof}

\subsubsection{Bounding $\mathtt{Tail}$}
The term $\mathtt{Tail}$ admits a direct bound, as stated in the following proposition.
\begin{proposition}[Bound on $\mathtt{Tail}$]\label{prop:tail}
	Under Assumptions~\ref{assump:1} and~\ref{assump:2}, the term $\mathtt{Tail}$ satisfies the upper bound
	\begin{align}
		\mathtt{Tail} \le K_n \Delta \tau_n |\mathcal{W}_n| B.
	\end{align}
\end{proposition}
\begin{IEEEproof}
	The result follows from the Cauchy-Schwarz inequality and Assumptions~\ref{assump:1} and~\ref{assump:2}.
\end{IEEEproof}

\section{Experiments}\label{sec:6}
In this section, we empirically evaluate the performance of the proposed DDAM-TOGD using a synthetic dataset and a real wireless traffic dataset \cite{sone2020wireless}.

\subsection{Baselines}\label{Sec:VI-A}
We compare the proposed DDAM-TOGD against the following three baselines.

\subsubsection{OGD (Sec.~\ref{Sec:II-A})} 
OGD is a centralized full-information scheme in which every agent has direct, delay-free access to all other agents' data. 
OGD is included as an \emph{oracle}-based lower-bound on the regret achievable by any decentralized protocol.

\subsubsection{C-DOGD (Sec.~\ref{Sec:II-B})} 
In C-DOGD, agents exchange and average iterates with their physical neighbors. 
As discussed in Sec.~\ref{Sec:II-B}, C-DOGD admits sublinear static regret only in the special case $\mathbf{W} = \mathbbm{1}\mathbbm{1}^T / N$, where all agents share a common memory objective. 
We nevertheless include it as the conventional distributed online learning baseline in order to quantify the performance gap that arises when the consensus assumption is violated.

\subsubsection{Truncated distributed OGD (Truncated-DOGD)} 
We further include a variant of DDAM-TOGD in which each agent communicates only with agents that are both physical neighbors and logically relevant according to the logical weight matrix $\mathbf{W}$. 
Any logically relevant but non-neighboring agents are simply ignored, so that Truncated-DOGD effectively solves a truncated version of the DDAM objective \eqref{eq:1_loss} in which the weight matrix is masked by the physical adjacency matrix. 
Truncated-DOGD reduces communication delay to a single hop, but it does not solve the original DDAM problem \eqref{eq:1_loss}. 
It is hence included to demonstrate the importance of tree-based communication in the proposed DDAM-TOGD is essential.

\vspace{-1em}
\subsection{Synthetic Dataset}
We consider memorization under the DeltaNet model \cite{schlag2021linear}, where memory retrieval is linear and the loss function for agent $n$ at time $t$ is given by the third entry in Table~\ref{tab:loss}.
The key vectors $\mathbf{k}_{n,t}$ are generated independently for each agent $n$ and time $t$, by drawing each entry from a uniform distribution over the interval $[-1, 1]$.
The corresponding value vectors $\mathbf{v}_{n,t}$ are generated as
\begin{equation}\label{eq:val_gen}
	\mathbf{v}_{n,t} = \left( (1 - \rho) \mathbf{M}_{n}^* + \rho \mathbf{M}_\mathrm{com}^* \right) \mathbf{k}_{n,t}  + \mathbf{n}_{n,t} , 
\end{equation}
where $\mathbf{n}_{n,t} \sim \mathcal{N}(0, \sigma_n^2)$ denotes additive Gaussian noise, with $\sigma_n^2 = 1$.
The data-generation model \eqref{eq:val_gen} indicates that, apart from the presence of noise, the ground-truth optimal linear mechanism for agent $n$ is given by
$ (1 - \rho)\mathbf{M}_n^* + \rho \mathbf{M}_\mathrm{com}^* $, combining a personalized component $ \mathbf{M}_n^* $ and a common component $ \mathbf{M}_\mathrm{com}^* $ shared across all agents. 
The correlation parameter \( \rho \in [0, 1] \) controls the trade-off between personalized and common memory contributions.
Matrix $ \mathbf{M}_\mathrm{com}^* $ is generated by sampling each entry independently from a chi-squared distribution with 2 degrees of freedom, while the matrices $ \mathbf{M}_n^* $ are generated independently with Gaussian entries $ \mathbf{M}_{n}^* \sim \mathcal{N}(\mu_n, \sigma_n^2) $, where the mean $\mu_n$ and variance $\sigma_n^2$ are independently and uniformly sampled from the intervals $[-5, 5]$ and $[0, 50]$, respectively.

\begin{figure}[t]
	\centering
	\includegraphics[width=0.765\linewidth]{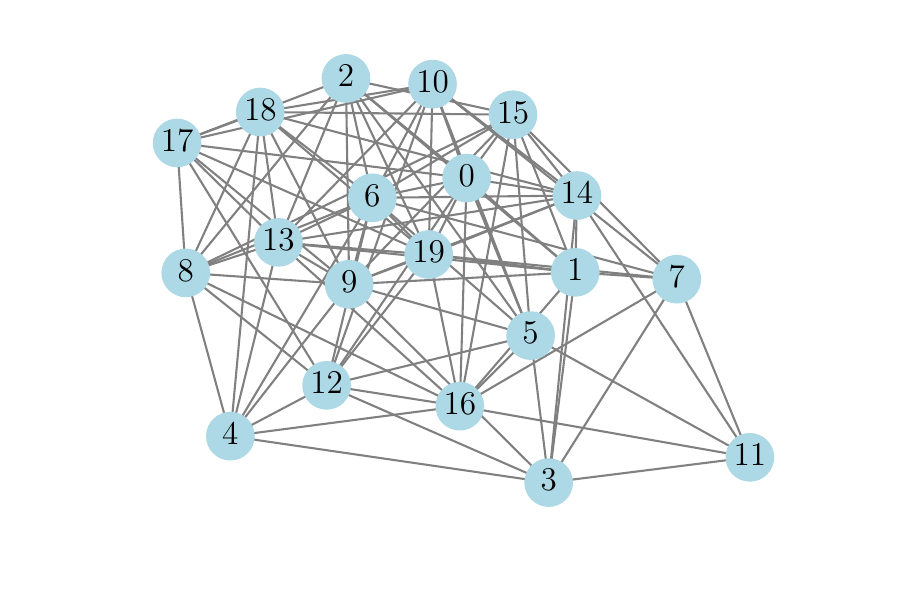}
	\vspace{-1em}
	\caption{Physical topology considered in the experiments.}
	\label{fig:exp1_tree}
\end{figure}

\begin{figure}[t]
	\centering
	\includegraphics[width=0.8\linewidth]{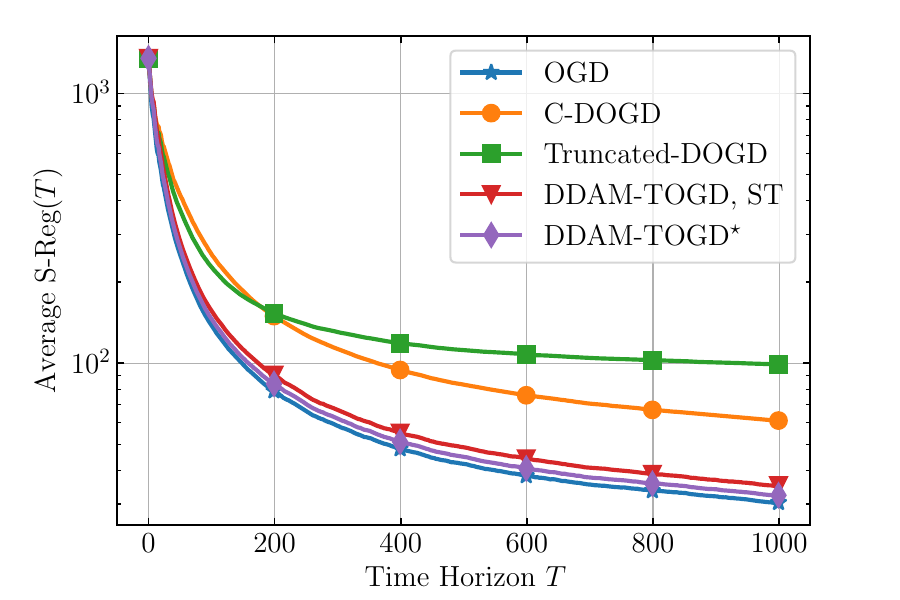}
	\vspace{-1em}
	\caption{Regret versus time horizon $T$ ($\rho = 0.75$, $y_0 = 2$, and $y_1 = 10$).}
	\label{fig:exp1_time}
\end{figure}

The logical weight matrix $\mathbf{W}$ is constructed such that each $n$-th row is sampled independently from a Dirichlet distribution.
Specifically, the $n$-th row is determined as the random vector
\begin{equation}\label{eq:24}
	\mathbf{W}[n , :] \sim \mathrm{Dirichlet} (  \underbrace{y_0 , \ldots y_0}_{1 : n-1} , y_1 , \underbrace{y_0 , \ldots y_0}_{n+1 : N} ) ,
\end{equation}
for some $y_1 \ge y_0 \ge 0$. 
This way, the weight in \eqref{eq:24} assigned to the local data at agent $ n $ is, on average,  more relevant than the data from other agents by a factor $ y_1 / y_0 $.
The number of agents is set to $N = 20$, and the connecting graph is shown in Fig. \ref{fig:exp1_tree}.

To ensure a fair comparison, given a time horizon $T$ for OGD and C-DOGD, we set the time horizon for DDAM-TOGD as $T / C_{\max}^{\mathcal{T}}$, where the maximum per-link capacity $C_{\max}^{\mathcal{T}}$ for a given choice of trees $\mathcal{T}$ is defined in \eqref{eq:capacity_max}.
This accounts for the fact that a single iteration may take up to $1/C_{\max}^{\mathcal{T}}$ more time in DDAM-TOGD as compared to C-DOGD, since the link with the largest load must communicate $C_{\max}^{\mathcal{T}}$ messages per iteration.
For DDAM-TOGD, we consider a baseline version using Steiner trees, referred to as ``DDAM-TOGD, ST'', with DDAM-TOGD$^{\star}$ defined as in Sec.~\ref{Sec-III-C}.
Recall that Steiner trees optimize the total edge weight, while the design introduced in Sec.~\ref{Sec-III-C} minimizes the total path-length.
The average static regret of the network at time horizon $T$ is $\mathrm{S}\text{-}\mathrm{Reg}(T) / N T$.

\noindent\textbf{Convergence:}
In Fig.~\ref{fig:exp1_time}, we study the evolution of the average static regret \eqref{eq:S_Reg} over time for OGD, C-DOGD, Truncated-DOGD, and the proposed DDAM-TOGD. 
As evidenced by the consistent decline in regret with increasing time horizon $T$, both OGD and the proposed DDAM-TOGD exhibit a clear sublinear regret trend, conforming to Theorem \ref{Theo:DAM-TOGD} and Corollary \ref{Coro:D_Reg_Conv}. 
In contrast, the regret of both C-DOGD and Truncated-DOGD initially decreases but quickly plateaus, exhibiting a regret behavior that fails to improve with time. 
For C-DOGD, this is because C-DOGD forces the agents to use the same memory mechanism as time $t$ increases, while the model \eqref{eq:val_gen} implies that different memorization mechanisms are preferred at different agents.
For Truncated-DOGD, each agent optimizes only over its physically adjacent and logically relevant agents, overlooking all logically relevant but non-neighboring agents, so that it converges to a truncated version of the memory, thus incurring an even higher regret than C-DOGD.

\begin{figure}[t]
	\centering
	\includegraphics[width=0.8\linewidth]{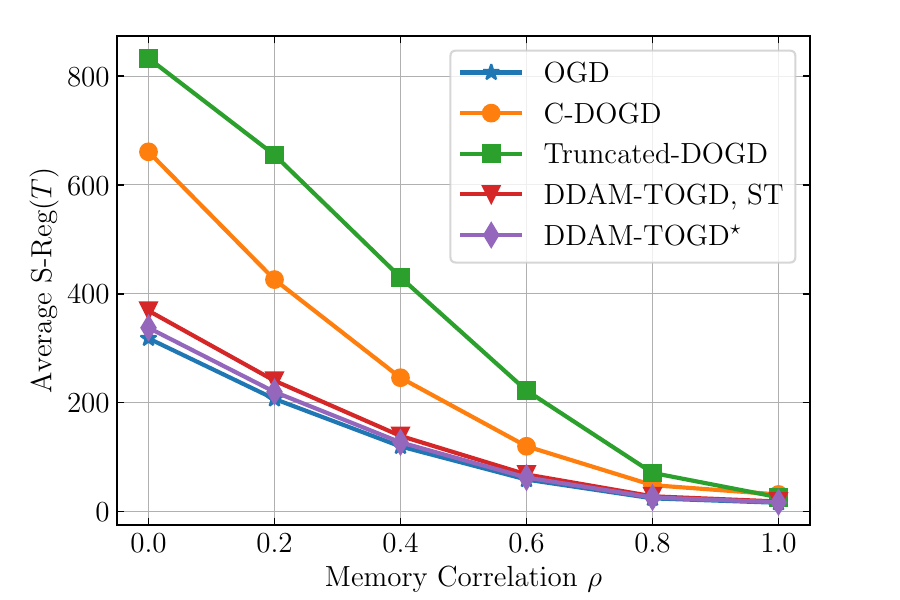}
	\vspace{-1em}
	\caption{Regret at $T = 1000$ versus memory correlation parameter $\rho$ ($y_0 = 2$, and $y_1 = 10$).}
	\label{fig:exp1_memory}
\end{figure}

\noindent\textbf{Impact of Memory Correlation:}
Fig.~\ref{fig:exp1_memory} shows the effect of the memory correlation parameter $\rho$ in \eqref{eq:val_gen} on the regret. 
For large $\rho$, memory heterogeneity is minimal and all methods perform similarly. 
As $\rho$ decreases, personalization increases and consensus becomes restrictive, causing the regret of both C-DOGD and Truncated-DOGD to grow rapidly. 
Truncated-DOGD degrades the most, since each agent ignores its logically relevant but non-neighboring agents. 
In contrast, DDAM-TOGD mitigates heterogeneity and maintains low regret, demonstrating robustness to personalization.

\begin{figure}[t]
	\centering
	\includegraphics[width=0.8\linewidth]{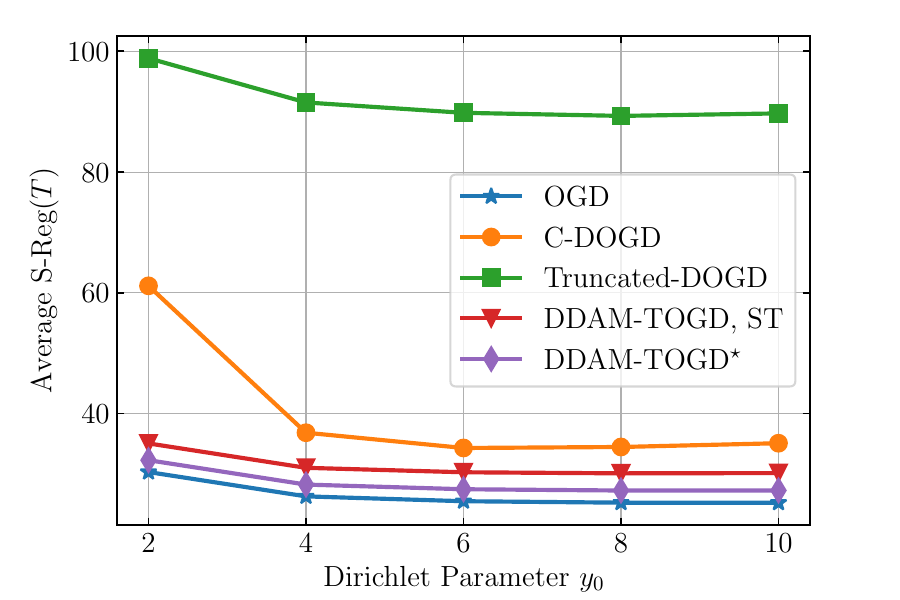}
	\vspace{-1em}
	\caption{Regret at $T = 1000$ versus Dirichlet parameter $y_0$ ($\rho = 0.75$, and $y_1 = 10$).}
	\label{fig:exp1_weights}
\end{figure}

\noindent\textbf{Impact of the Logical Weight Matrix:}
In Fig.~\ref{fig:exp1_weights}, we report the effect of the logical weight matrix in \eqref{eq:24} by varying the Dirichlet parameter $y_0$ (with $y_1=10$), which controls how each agent’s memory incorporates data from others.
When the parameter $y_0$ is small, the weight matrix $\mathbf{W}$ becomes highly imbalanced, emphasizing personalized weights. 
In this regime, achieving consensus does not result in the optimal solution and C-DOGD incurs a large regret.
Truncated-DOGD, in contrast, incurs a large regret across the entire range of $y_0$, since restricting communication to physically adjacent agents discards logically relevant information regardless of how the logical weights are distributed. 
The results show that OGD and the proposed DDAM-TOGD methods are largely unaffected by variations in parameter $y_0$, demonstrating their robustness to changes in the network logical weight structure.

\vspace{-1em}
\subsection{Wireless Traffic Dataset}

\begin{figure}[t]
	\centering
	\includegraphics[width=0.765\linewidth]{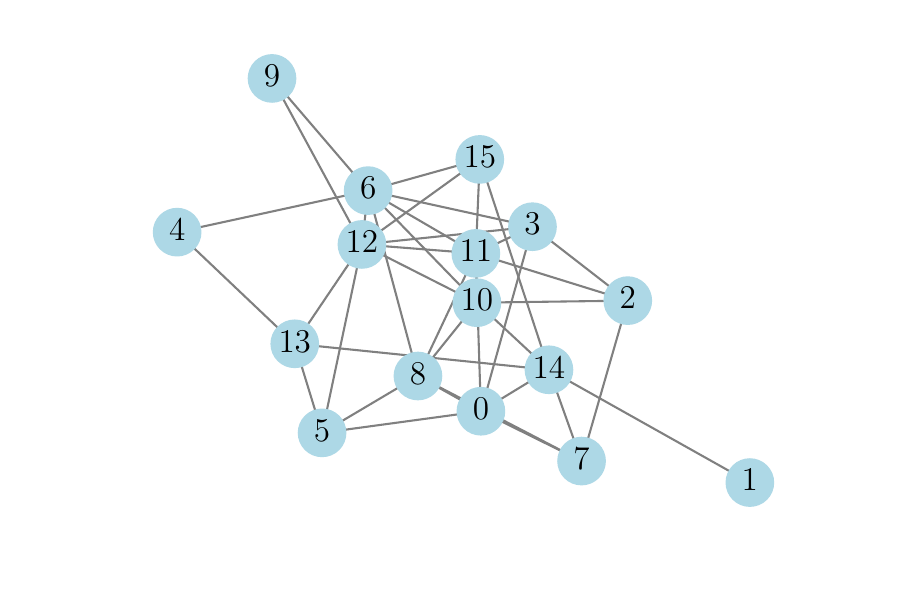}
	\vspace{-1em}
	\caption{Physical topology considered in the experiments with the wireless traffic dataset.}
	\label{fig:exp2_tree}
\end{figure}

In this experiment, we evaluate wireless traffic usage memorization using real traffic data from an enterprise network~\cite{sone2020wireless}. 
The dataset consists of measurements collected from $470$ access points (APs). 
In our experiment, we focus on $N=16$ APs randomly selected from the ``High'' group \cite{sone2020wireless}, which consists of 79 APs with the highest mean traffic usage.
To reduce the dynamic range of inputs, raw traffic measurements are preprocessed using a logarithmic transformation.
In this experiment, the connection graph is randomly generated, where each pair of nodes is connected with probability $0.25$. 
The resulting graph is illustrated in Fig.~\ref{fig:exp2_tree}.

In this dataset, the transmitted traffic volume was recorded every 10 minutes between December 20, 2018 and February 8, 2019 \cite{sone2020wireless}.
Thus, each hour consists of six consecutive samples for every AP. 
For each AP $n$ and time step $t$, we form a key–value pair $(\mathbf{k}_{n,t}, \mathbf{v}_{n,t})$, where the value $\mathbf{v}_{n,t} \in \mathbb{R}^{d_v}$ aggregates the six ($d_v = 6$) 10-minute traffic measurements of AP $n$ within the hour associated with time step $t$.
The corresponding key is constructed as
$ \mathbf{k}_{n,t} = 
[\mathbf{k}_{\mathrm{AP},n}; \; \mathbf{k}_{\mathrm{time},t}]
\in \mathbb{R}^{d_k} $, 
where $\mathbf{k}_{\mathrm{AP},n} = \Phi_{\mathrm{AP}} (n) \in \mathbb{R}^{d_{\mathrm{AP}}}$ specifies the identity of the $n$-th AP, and 
$\mathbf{k}_{\mathrm{time},t} = \Phi_{\mathrm{time}} (t\mod 24) \in \mathbb{R}^{d_{\mathrm{time}}}$ encodes the hour-of-day associated with time step $t$. 
The temporal embedding function $\Phi_{\mathrm{time}}$ uses Transformer-style sinusoidal (cos/sin) embeddings \cite{vaswani2017attention}, whereas the AP embedding function $\Phi_{\mathrm{AP}}$ combines a one-hot representation of the AP index with a Transformer-style sinusoidal embedding.
In our experiments, we set $d_{\mathrm{AP}} = 24$, and $d_{\mathrm{time}} = 10$.

To evaluate the dynamic regret, we use as a reference a comparator $\mathbf{U}_{n,t}$ that is optimized in hindsight separately, for each time window
$ \Theta_k \triangleq \{(k-1)\Omega+1,\ldots,k\Omega\}$ with $k=1,2,\ldots $, of duration $\Omega$ time intervals, i.e., $\Omega \times 10$ minutes.
Accordingly, we select the comparator $\mathbf{U}_{n,t}$, for all $t\in\Theta_k$, as
\begin{equation}
	\label{eq:windowed-comparator}
	\mathbf{U}_{n,t} \in \arg\min_{\mathbf{U}\in\mathcal{X}} 
	\sum_{s\in\Theta_k}\ \sum_{m\in\mathcal{W}_n} w_{n,m}\, f_{m,s}\big(\mathbf{U}\big),
\end{equation}
If $\Omega = 1$, the comparator \eqref{eq:windowed-comparator} can change at every time step, while if $\Omega$ equals the time horizon $T$, the comparator \eqref{eq:windowed-comparator} reduces to the hindsight-optimal static solution in \eqref{eq:best_solution}, yielding the static-regret criterion.

The time-horizon scaling method and logical weights are set in the same way as in the synthetic-dataset case described in the previous subsection.
The average dynamic regret of network at time horizon $T$ is evaluated as the ratio $\mathrm{D}\text{-}\mathrm{Reg}(T) / NT$.

\begin{figure}[t]
	\centering
	\includegraphics[width=0.8\linewidth]{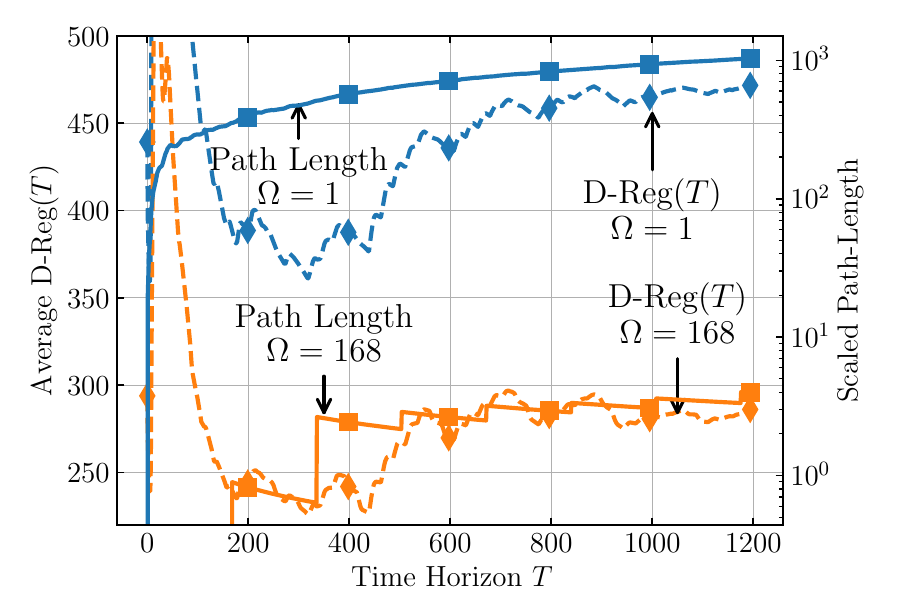}
	\vspace{-1em}
	\caption{Regret of proposed DDAM-TOGD$^\star$ and scaled path-length versus time horizon $T$ with different $\Omega$  ($y_0 = 10$, and $y_1 = 100$).}
	\label{fig:exp2_PL}
\end{figure}
\noindent\textbf{Validation of the relation between dynamic regret and path-length:}
We plot the dynamic regret for the proposed DDAM-TOGD$^\star$ with different time window size $\Omega$ in Fig.~\ref{fig:exp2_PL}. 
To validate the bounds \eqref{eq:DAM-TOGD-Regret-1}, we also show the quantity $\sum_{n \in \mathcal{N}}(1 + \mathrm{PL}_n^T) / N \sqrt{T}$, which describes the scaling of the bounds on the time-averaged dynamic regret.
When $\Omega = 1$, the comparator may change at every time step, causing the path-length to increase rapidly. 
In contrast, when $\Omega = 168$, the comparator changes every week (since $\Omega = 7 \times 24$), and the path-length grows much more slowly. 
Validating the theoretical bounds, in the case $\Omega = 1$, the dynamic regret also exhibits a fast-increasing trend, whereas for $\Omega = 168$ it grows significantly more slowly. 
Overall, the dynamic regret closely follows the evolution of the scaled path-length.

\begin{figure}[t]
	\centering
	\includegraphics[width=0.8\linewidth]{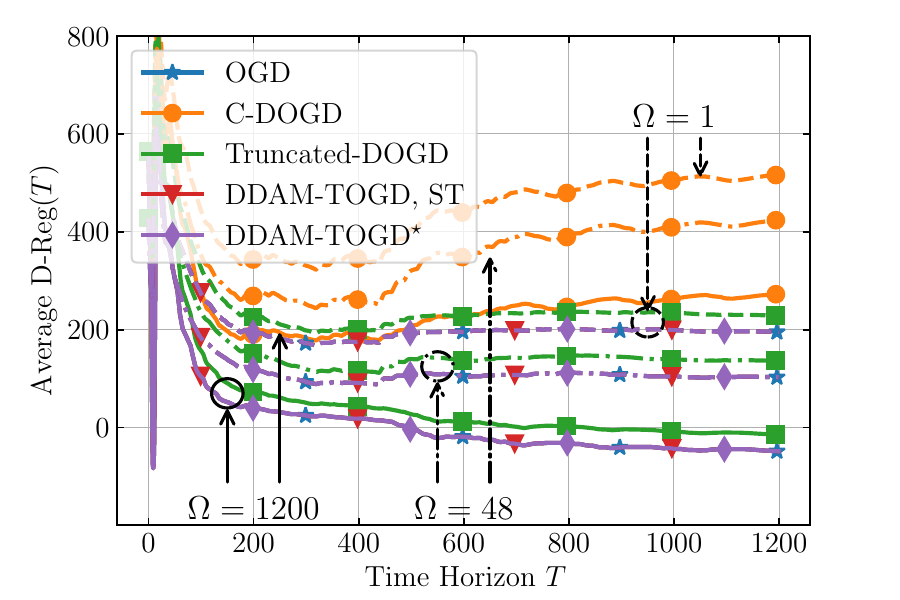}
	\vspace{-1em}
	\caption{Regret versus time horizon $T$ with different $\Omega$  ($y_0 = 0.1$, and $y_1 = 100$). 
	}
	\label{fig:exp2_time}
\end{figure}

\noindent\textbf{Convergence:}
In Fig. \ref{fig:exp2_time}, we study the evolution of the dynamic regret \eqref{eq:D_Reg} over time for OGD, C-DOGD, Truncated-DOGD, and the proposed DDAM-TOGD with different choices of the time window size $\Omega$ of the comparators \eqref{eq:windowed-comparator}. 
Different colors and markers denote distinct methods, whereas different line styles denote different values of $\Omega$.
The proposed DDAM-TOGD is seen to approach the ideal centralized solution set by OGD.
This shows that in this setting communication delays have a minor impact on performance due to the near-periodicity of the data. 
In contrast, for C-DOGD, the dynamic regret is much larger than for OGD, indicating the need for specializing the AM mechanisms at the agents.
Interestingly, Truncated-DOGD performs worse than OGD and the proposed method, but better than C-DOGD. 
This is because $y_0=0.1$ and $y_1=100$ make each agent emphasize its own data, so that the truncated AM closely approximates the optimal AM.
Note also that the static regret, obtained with $\Omega = T = 1200$, can become negative, showing that the hindsight-best solution is not necessarily optimal in a non-stationary environment.

\begin{figure}[t]
	\centering
	\includegraphics[width=0.8\linewidth]{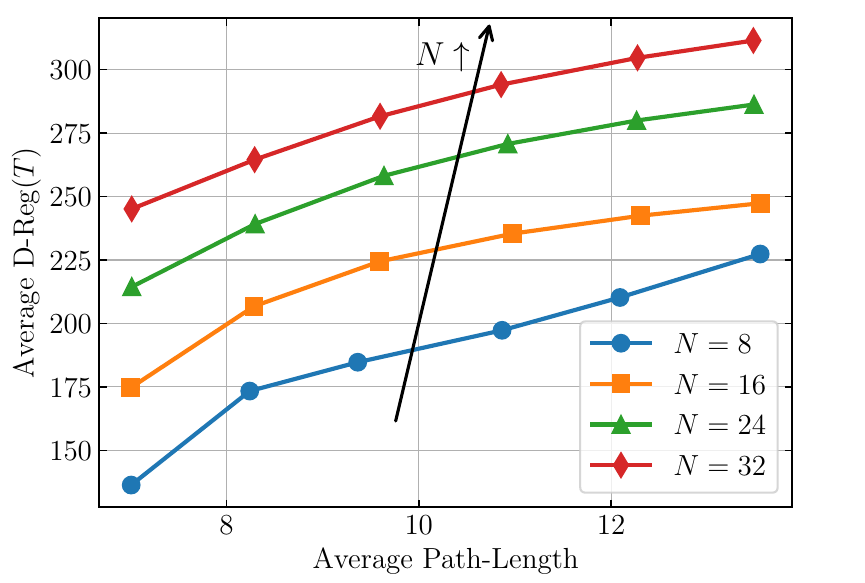}
	\vspace{-1em}
	\caption{Regret of proposed DDAM-TOGD$^\star$ at $T = 1200$ versus average path-length ($y_0 = 10$, $y_1 = 100$, and $\Omega = 1$).}
	\label{fig:exp2_DReg_PathLen}
\end{figure}

\noindent\textbf{Impact of Network Size and Data Variability:}
In Fig.~\ref{fig:exp2_DReg_PathLen}, we investigate the impact of the network size $N$ on the dynamic regret.
To ensure a fair comparison across different network sizes, we apply low-pass filtering to the data with varying degrees of smoothing, so that the average path-length, i.e., $\sum_{n \in \mathcal{N}} \mathrm{PL}_n^T / NT$, is kept approximately the same across all network sizes.
A larger average path-length corresponds to more rapidly varying data, and hence reflects stronger data variability.
The results show that the dynamic regret of the proposed DDAM-TOGD$^\star$ grows with the path-length, consistent with the path-length dependence of the regret bound.
Moreover, for a fixed path-length, the dynamic regret increases with the network size $N$.
This is because a larger network increases both the cumulative communication delay $\tau_{n,\mathrm{sum}}$ and the maximum delay $\tau_{n,\max}$, which enlarges the regret bound.
These results are consistent with the theoretical analysis in Theorem~\ref{Theo:DAM-TOGD} and Corollary~\ref{Coro:D_Reg_Conv}.

\begin{figure}[t]
	\centering
	\includegraphics[width=0.8\linewidth]{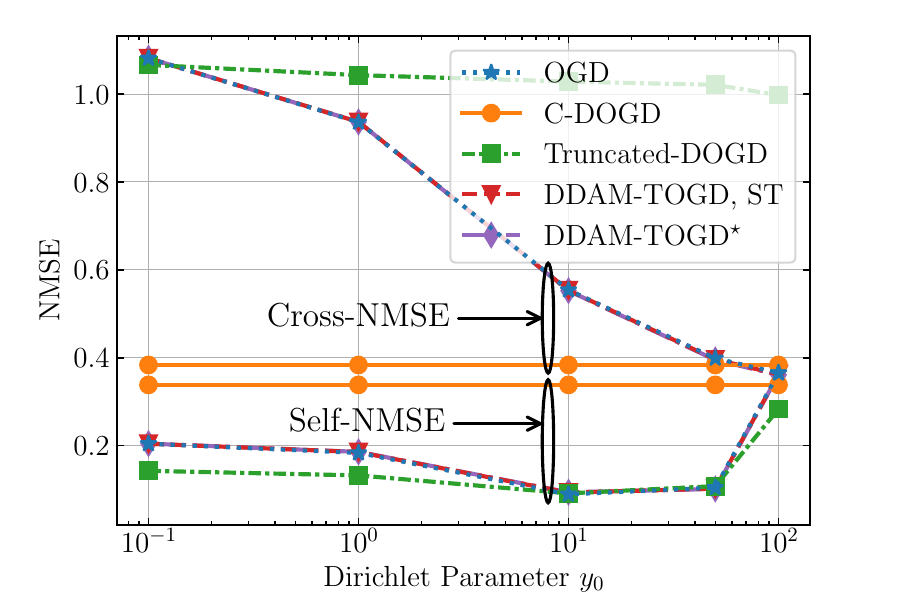}
	\vspace{-1em}
	\caption{NMSE versus Dirichlet parameter $y_0$ ($y_1 = 100$).}
	\label{fig:exp2_NMSE}
\end{figure}
\noindent\textbf{Normalized Mean Squared Error (NMSE) Performance:}
In addition to the regret performance shown so far, in Fig.~\ref{fig:exp2_NMSE} we investigate the NMSE performance of memory retrieval. 
Specifically, we evaluate self-NMSE and cross-NMSE, measuring the error in retrieving one’s own value and the ability to recall other APs’ traffic usage, respectively, defined as
\begin{align}
	\mathrm{Self}\text{-}\mathrm{NMSE} = &  \frac{\sum_t^T \sum_{n \in \mathcal{N}} \| \mathbf{X}_n \mathbf{k}_{n , t} - \mathbf{v}_{n , t} \|^2}{\sum_t^T \sum_{n \in \mathcal{N}} \| \mathbf{v}_{n , t} \|^2}  , \notag \\
	\mathrm{Cross}\text{-}\mathrm{NMSE} = & \frac{\sum_t^T \sum_{n \in \mathcal{N}} \sum_{m \in \mathcal{W}_n \backslash \{n\} } \| \mathbf{X}_n \mathbf{k}_{m , t} - \mathbf{v}_{m , t} \|^2}{\sum_t^T \sum_{n \in \mathcal{N}} \sum_{m \in \mathcal{W}_n \backslash \{n\} } \| \mathbf{v}_{m , t} \|^2} . \notag
\end{align}
In Fig. \ref{fig:exp2_NMSE}, as $y_0$ increases, each agent places less weight on its own data and more weight on other APs, which enhances its ability to memorize the traffic usage of other APs leading to a decreasing trend in cross-NMSE.
Furthermore, as $y_0$ increases, the self-NMSE first decreases slightly and then rises sharply. 
This observation suggests that selectively incorporating information from others can be beneficial, but excessive attention to other agents' data is detrimental.
In contrast, C-DOGD remains insensitive to $y_0$, with both its self- and cross-NMSE staying nearly constant, since it achieves consensus memory mechanisms.
Truncated-DOGD attains a low self-NMSE comparable to DDAM-TOGD but a persistently high cross-NMSE, as it always pays high attention to its own data but ignores all non-neighboring agents, regardless of the logical weights.

\section{Conclusion}\label{sec:7}
This paper introduced the concept of DDAM and proposed DDAM-TOGD, a tree-based distributed online gradient optimization method that enables agents to build and update AMs under time-varying data streams. 
We established sublinear static regret and path-length dependent dynamic regret, revealing how communication delay and routing topology affect regret performance.
Based on these insights from the theoretical analysis, we proposed a combinatorial optimization based routing-tree design method.
Experiments on synthetic datasets and real wireless traffic usage memorization have verified the theory: DDAM-TOGD adapts to changing associations, and achieves lower regret than existing solutions.

Future work includes investigating time-varying connectivity, exploring non-linear associative memory mechanisms, studying parameter-free online learning algorithms \cite{orabona2016coin} and studying memory-efficient algorithms that achieve tighter dynamic regret bounds \cite{zhang2018adaptive}.

\appendices
\section{Proof of Lemma \ref{lem:single_step}}\label{App:lem:single_step}
\renewcommand{\theequation}{A-\arabic{equation}}
\setcounter{equation}{0}

We first present the following inequality
\begin{align}
	& \frac{1}{2} \left\| \mathbf{X}_{n , t+1} - \mathbf{U}_{n,t} \right\|^2 - \frac{1}{2} \left\| \mathbf{X}_{n , t} - \mathbf{U}_{n,t} \right\|^2 \notag \\
	& = \frac{1}{2} \left\| \Pi_{\mathcal{X}} \left[  \mathbf{X}_{n , t} \! -  \! \eta_{n} \!\!\! \sum_{m \in \mathcal{W}_n} \!\!\! w_{n , m} \mathbf{G}_{m , t - \tau_{n , m}}^n \mathbbm{1}_{\{t > \tau_{n , m}\}} \right] \! \!- \!  \mathbf{U}_{n,t} \right\|^2 \notag \\
	& \quad - \frac{1}{2} \left\| \mathbf{X}_{n , t} - \mathbf{U}_{n,t} \right\|^2 \notag \\
	& \mathop \le \limits^{\text{(a)}} \frac{1}{2} \left\| \mathbf{X}_{n , t} - \eta_{n} \sum_{m \in \mathcal{W}_n} w_{n , m} \mathbf{G}_{m , t - \tau_{n , m}}^n \mathbbm{1}_{\{t > \tau_{n , m}\}} - \mathbf{U}_{n,t} \right\|^2 \notag \\
	& \quad - \frac{1}{2} \left\| \mathbf{X}_{n , t} - \mathbf{U}_{n,t} \right\|^2 \notag \\
	& = \frac{\eta_{n}^2}{2} \left\| \sum_{m \in \mathcal{W}_n} w_{n , m} \mathbf{G}_{m , t - \tau_{n , m}}^n \mathbbm{1}_{\{t > \tau_{n , m}\}} \right\|^2 \notag \\
	& \quad - \eta_{n} \sum_{m \in \mathcal{W}_n} \left\langle w_{n , m} \mathbf{G}_{m , t-\tau_{n , m}}^n \mathbbm{1}_{\{t > \tau_{n , m}\}} , \mathbf{X}_{n , t} - \mathbf{U}_{n,t} \right\rangle , \label{eq:App_A_2}
\end{align}
where (a) holds since $\left\| \Pi_{\mathcal{X}} [ \mathbf{X} ] - \mathbf{Y}  \right\|^2 \le \left\| \mathbf{X} - \mathbf{Y}  \right\|^2$ for $ \mathbf{Y}\in\mathcal{X}$ \cite[Proposition 2.11]{orabona2019modern}.

Dividing both sides by $\eta_{n}$ and rearranging the terms, we obtain
\begin{align}
	& \sum_{m \in \mathcal{W}_n} \left\langle w_{n , m} \mathbf{G}_{m , t-\tau_{n , m}}^n \mathbbm{1}_{\{t > \tau_{n , m}\}} , \mathbf{X}_{n , t} - \mathbf{U}_{n,t} \right\rangle  \notag \\
	& \le \frac{\eta_{n}}{2} \left\| \sum_{m \in \mathcal{W}_n} w_{n , m} \mathbf{G}_{m , t - \tau_{n , m}}^n \mathbbm{1}_{\{t > \tau_{n , m}\}} \right\|^2 \notag \\
	& \quad + \frac{\left\| \mathbf{X}_{n , t} - \mathbf{U}_{n,t} \right\|^2 - \left\| \mathbf{X}_{n , t+1} - \mathbf{U}_{n,t} \right\|^2}{2 \eta_{n}} \notag \\
	& \mathop \le \limits^{\text{(a)}} \frac{\eta_{n}}{2} \sum_{m \in \mathcal{W}_n} w_{n , m}  \left\| \mathbf{G}_{m , t - \tau_{n , m}}^n \right\|^2 \notag \\
	& \quad + \frac{\left\| \mathbf{X}_{n , t} - \mathbf{U}_{n,t} \right\|^2 - \left\| \mathbf{X}_{n , t+1} - \mathbf{U}_{n,t} \right\|^2}{2 \eta_{n}} \notag \\
	& \mathop \le \limits^{\text{(b)}} \left( \sum_{m \in \mathcal{W}_n} w_{n,m} G_m^2 \right)  \frac{\eta_{n}}{2} \notag \\
	& \quad + \frac{\left\| \mathbf{X}_{n , t} - \mathbf{U}_{n,t} \right\|^2 - \left\| \mathbf{X}_{n , t+1} - \mathbf{U}_{n,t} \right\|^2}{2 \eta_{n}} \notag \\
	& \mathop \le \limits^{\text{(c)}} K_n \left( \sum_{m \in \mathcal{W}_n}  G_m \right)  \frac{\eta_{n}}{2}  \notag \\
	& \quad + \frac{\left\| \mathbf{X}_{n , t} - \mathbf{U}_{n,t} \right \|^2 - \left\| \mathbf{X}_{n , t+1} - \mathbf{U}_{n,t} \right\|^2}{2 \eta_{n}}, \label{eq:App_A_3_1}
\end{align}
where (a) follows from Jensen's inequality, (b) from the bounded gradient assumption, and (c) from the definition of $K_n$.

\section{Proof of Lemma \ref{lem:telescoping}}\label{App:lem:telescoping}
\renewcommand{\theequation}{B-\arabic{equation}}
\setcounter{equation}{0}
It follows
\begin{align}
		& \sum_{t=\tau_{n , \min} +1}^{T + \tau_{n , \max}} \frac{\left\| \mathbf{X}_{n , t} - \mathbf{U}_{n , t} \right\|^2 - \left\| \mathbf{X}_{n , t+1} - \mathbf{U}_{n , t} \right\|^2}{2 \eta_{n}} \notag \\
		&  = \sum_{t=\tau_{n , \min} +1}^{T + \tau_{n , \max}} \frac{\left\| \mathbf{X}_{n , t} \right\|^2 - \left\| \mathbf{X}_{n , t+1} \right\|^2}{2 \eta_{n}} \notag \\
		& \quad + \sum_{t=\tau_{n , \min} +1}^{T + \tau_{n , \max}} \frac{ \left\langle \mathbf{U}_{n,t} , \mathbf{X}_{n , t+1} \right\rangle  - \left\langle  \mathbf{U}_{n,t} , \mathbf{X}_{n , t} \right\rangle }{\eta_{n}} \label{eq:app:B-1}
\end{align}

For the first term in \eqref{eq:app:B-1}, we have
\begin{align}
	& \sum_{t=\tau_{n , \min} +1}^{T+\tau_{n,\max}} \frac{\left\| \mathbf{X}_{n , t} \right\|^2 - \left\| \mathbf{X}_{n , t+1} \right\|^2}{2 \eta_{n}} \notag \\
	& = \frac{\left\|\mathbf{X}_{n , \tau_{n,\min}+1}\right\|^2}{2\eta_{n}} 
	- \frac{\left\|\mathbf{X}_{n , T+{\tau_{n,\max+1}}}\right\|^2}{2\eta_{n}} \notag \\
	& \le \frac{B^2}{2\eta_{n}} .
\end{align}

For the second term in \eqref{eq:app:B-1}, we have
\begin{align}
	& \sum_{t=\tau_{n , \min} +1}^{T+\tau_{n,\max}} \frac{ \left\langle \mathbf{U}_{n,t} , \mathbf{X}_{n , t+1} \right\rangle  - \left\langle  \mathbf{U}_{n,t} , \mathbf{X}_{n , t} \right\rangle }{\eta_{n}} \notag \\
	& = \sum_{t=\tau_{n , \min} + 2}^{T+\tau_{n,\max}} \frac{ \left\langle \mathbf{ U }_{n,t-1} - \mathbf{U}_{n,t} , \mathbf{ X }_{n,t} \right\rangle  }{\eta_{n}} \notag \\
	& \quad + \frac{ \left\langle \mathbf{ U }_{n,T+\tau_{n,\max}} , \mathbf{ X }_{n,T+\tau_{n,\max}+1} \right\rangle  }{\eta_{n}}  \notag \\
	& \quad - \frac{\left\langle \mathbf{ U }_{n,\tau_{n,\min}+1} , \mathbf{ X }_{n,\tau_{n,\min}+1} \right\rangle}{\eta_n} \notag \\
	& \le B  \frac{ \sum_{t=\tau_{n , \min} + 2}^{T+\tau_{n,\max}} \left\| \mathbf{ U }_{n,t-1} - \mathbf{U}_{n,t}\right\|  }{\eta_{n}} + \frac{B^2}{4 \eta_n} + \frac{B^2}{\eta_n} \notag \\
	& \le \frac{B}{\eta_n} \mathrm{PL}_n^{T+\tau_{n,\max}} + \frac{5B^2}{4\eta_n} , \label{Neq:C2-5}
\end{align}
with the first inequality in \eqref{Neq:C2-5} following from $- B^2 / 4 \le \left\langle \mathbf{X} , \mathbf{Y} \right\rangle \le B^2$ \cite[Appendix A]{zinkevich2003online}.

Combining the above results completes the proof.

\section{Proof of Bound on $\mathtt{Drift}_{\mathbf{X}}$}\label{App:eq:reg_drift_2}
\renewcommand{\theequation}{C-\arabic{equation}}
\setcounter{equation}{0}
Unrolling the difference $\mathbf{X}_{n,t} - \mathbf{X}_{n,t-\tau_{n,m}}$ for $t > \tau_{n , m}$ and from the update rule \eqref{eq:6}, it holds
\begin{align}
	& \left\| \mathbf{X}_{n , t} - \mathbf{X}_{n , t -\tau_{n , m}} \right\| \\
	& = \left\| \sum_{j=1}^{\tau_{n , m}} ( \mathbf{X}_{n , t-j+1} - \mathbf{X}_{n , t-j} ) \right\| \notag \\
	& \mathop \le \limits^{\text{(a)}} \sum_{j=1}^{\tau_{n , m}} \left\|  \mathbf{X}_{n , t-j+1} - \mathbf{X}_{n , t-j} \right\| \notag \\
	& \mathop \le \limits^{\text{(b)}} \eta_{n} \sum_{j=1}^{\tau_{n , m}}  \left\|  \sum_{s \in \mathcal{W}_n} w_{n , s} \mathbf{G}_{s , t-j-\tau_{n,s}}^n \mathbbm{1}_{ \{ t-j > \tau_{n,s} \} } \right\|  \notag \\
	& \mathop \le \limits^{\text{(c)}} \eta_{n} \sum_{j=1}^{\tau_{n , m}}    \sum_{s \in \mathcal{W}_n} w_{n , s} \left\| \mathbf{G}_{s , t-j-\tau_{n,s}}^n \right\| \mathbbm{1}_{ \{ t-j > \tau_{n,s} \} }   \notag \\
	& \mathop \le \limits^{\text{(d)}}  \eta_{n} \sum_{s \in \mathcal{W}_n} \sum_{j=1}^{ \min \{ \tau_{n , m} , t-\tau_{n , s}-1  \} }  w_{n , s} G_s  \notag \\
	& \mathop \le \limits^{\text{(e)}}  \eta_{n} K_n \sum_{s \in \mathcal{W}_n} \min \{ \tau_{n , m} , t-\tau_{n , s}-1  \} , \label{eq:reg_drift_2}
\end{align}
where (a) and (c) hold by the triangle inequality, 
(b) follows from $\left\| \Pi_{\mathcal{X}} [ \mathbf{X} ] - \mathbf{Y}  \right\|^2 \le \left\| \mathbf{X} - \mathbf{Y}  \right\|^2$ for $\mathbf{Y} \in\mathcal{X}$ \cite[Proposition 2.11]{orabona2019modern}, (d) holds by the bounded gradient assumption,
and (e) follows from $\sum_{j=1}^{\min\{ a , b\}} c = \min\{ a , b\} \times c$, with $c$ being a constant independent of $j$.

Substituting \eqref{eq:reg_drift_1} into \eqref{eq:reg_drift_2} gives
\begin{align}
	& \mathtt{Drift}_{\mathbf{X}} \notag \\
	& \le \sum_{m \in \mathcal{W}_n} K_n^2 \eta_{n}  \sum_{t=\tau_{n ,m}+1}^{T+\tau_{n, m}} \sum_{s \in \mathcal{W}_n}  \min\{ \tau_{n , m} , t-\tau_{n , s}-1 \}  \notag \\
	& \le \sum_{t=\tau_{n ,\min}+1}^{T+\tau_{n, \max}} \!\!\!\! K_n^2 \eta_{n} \!\!\!\!  \sum_{m \in \mathcal{W}_n}  \sum_{s \in \mathcal{W}_n}  \!\!\!\! \min\{ \tau_{n , m} , t\!-\!\tau_{n , s}\!-\!1 \} \mathbbm{1}_{\{t > \tau_{n , m}\}} \notag \\
	& = \sum_{m \in \mathcal{W}_n} \sum_{s \in \mathcal{W}_n}  \sum_{t = \tau_{n,m} + 1}^{ \tau_{n,s} + \tau_{n , m}} K_n^2 \eta_{n} ( t - \tau_{n,s} - 1 ) \notag \\
	& \quad + \sum_{m \in \mathcal{W}_n} \sum_{s \in \mathcal{W}_n} \sum_{ t = \tau_{n,s} + \tau_{n , m}  + 1 }^{ T + \tau_{n , \max} } K_n^2 \eta_{n}  \tau_{n , m}  \notag \\
	& = \sum_{m \in \mathcal{W}_n} \eta_{n}  K_n^2  \sum_{s \in \mathcal{W}_n} \sum_{t = \tau_{n,m} + 1}^{ \tau_{n,s} + \tau_{n , m} } ( 
	t - \tau_{n,s} - 1 ) \notag \\ \label{NeqC:p2-5}
	& \quad+ \sum_{m \in \mathcal{W}_n} \tau_{n , m} K_n^2 \sum_{s \in \mathcal{W}_n}  \sum_{ t = \tau_{n,s} + \tau_{n , m}  + 1 }^{ T + \tau_{n , \max} } \eta_{n} . 
\end{align}

For the first term in the right side of \eqref{NeqC:p2-5}, we have
\begin{equation}
	\begin{aligned}
		& \sum_{s \in \mathcal{W}_n} \sum_{t = \tau_{n,m} + 1}^{ \tau_{n,s} + \tau_{n , m} } ( 
		t - \tau_{n,s} - 1 ) \\
		& = \sum_{s \in \mathcal{W}_n} \frac{2 \tau_{n , m} - \tau_{n , s} - 1}{2} \tau_{n , s} \\
		& \le |\mathcal{W}_n| \tau_{n, \max}^2 .
	\end{aligned}
\end{equation}

For the second term in the right side of \eqref{NeqC:p2-5}, we have
\begin{equation}
	\begin{aligned}
		\sum_{s \in \mathcal{W}_n}  \sum_{ t = \tau_{n,s} + \tau_{n , m} + 1}^{ T + \tau_{n , \max} } \eta_{n} 
		& \le \sum_{s \in \mathcal{W}_n}  \sum_{ t = \tau_{n,\min} }^{ T + \tau_{n , \max} } \eta_{n} \\
		& = |\mathcal{W}_n| \sum_{t = \tau_{n,\min}+1}^{T + \tau_{n , \max}} \eta_{n} \\
		& = |\mathcal{W}_n| \eta_{n} \left( T + \Delta \tau_n \right).
	\end{aligned}
\end{equation}

Combining the above results, we complete this proof.

\section{Proof of Lemma \ref{lemm:drift_2}}\label{App:drift_2}
\renewcommand{\theequation}{D-\arabic{equation}}
\setcounter{equation}{0}
By unfolding the difference $\mathbf{U}_{n , t} - \mathbf{U}_{n,t-\tau_{n, m}}$, we obtain
\begin{align}
	& \sum_{t=\tau_{n ,m}+1}^{T+\tau_{n, m}} \left\| \mathbf{U}_{n , t} - \mathbf{U}_{n,t-\tau_{n, m}} \right\| \notag \\
	& = \sum_{t=\tau_{n ,m}+1}^{T+\tau_{n, m}}  \left\| \sum_{s = 1}^{ \tau_{n,m}} \left( \mathbf{U}_{n , t-s+1} - \mathbf{U}_{n,t-s} \right) \right\| \notag \\
	& \le \sum_{t=\tau_{n ,m}+1}^{T+\tau_{n, m}} \sum_{s = 1}^{ \tau_{n,m}} \left\| \mathbf{U}_{n , t-s+1} - \mathbf{U}_{n,t-s} \right\| \notag \\
	& = \sum_{t=2}^{T+\tau_{n, m}} \xi_{n,m}^t \left\| \mathbf{U}_{n , t} - \mathbf{U}_{n,t-1} \right\| ,
\end{align}
where
\begin{equation}
	\xi_{n,m}^t = \left\{
	\begin{array}{ll}
		t-1 , & 2 \le t \le \tau_{n,m} \\
		\tau_{n,m} , & \tau_{n,m} < t <T \\
		T+\tau_{n,m}-t+1 , & T \le t \le T+\tau_{n,m} \\
	\end{array}
	\right. .
\end{equation}
Since $\xi_{n,m}^t \le \tau_{n,m}$ for all $t$, we complete this proof.

\section{Combinatorial Optimization-based Tree Design}\label{App:Tree_Optim}
\renewcommand{\theequation}{E-\arabic{equation}}
\setcounter{equation}{0}

We formulate an optimization problem tailored to our specific design objectives. 
This formulation is grounded in a flow-conservation-based approach, which accurately models the connectivity requirements and the path-dependent cost characteristics described below:
\begin{subequations}\label{eq:steiner_conn_ilp}
	\begin{align}
		& \min_{\substack{
				h_{i , j} \in \{ 0 , 1 \},\; v_i \in \{ 0 , 1 \} \\
				C_{i , j}^w \ge 0,\;
				C^{\text{conn}}_{i , j} \ge 0
		}}  \mathrm{Dist} = \sum_{w \in \mathcal{W}_n} \sum_{(i,j) \in \mathcal{E}} \!\!\! \tau_{i , j} \left( C_{i , j}^w + C_{j , i}^w \right)
		\label{eq:steiner_obj} \\
		& \mathrm{s.t.} \;\;
		v_w = 1, \quad \forall w \in \mathcal{W}_n \label{eq:steiner_target_activated} \\
		& \quad \sum_{(i , j) \in \mathcal{E}} h_{i , j} = \sum_{i \in \mathcal{N}} v_i - 1 \label{eq:steiner_tree_structure} \\
		& \quad h_{i , j} \le v_i,\quad h_{i , j} \le v_j, \; \forall (i , j) \in \mathcal{E} \label{eq:steiner_edge_node_consistency} \\
		& \quad C_{i , j}^w + C_{j , i}^w \le h_{i , j}, \; \forall w \in \mathcal{W}_n \setminus \{n\},\; \forall (i , j) \in \mathcal{E} \label{eq:steiner_flow_edge_usage} \\
		& \quad \sum_{j : (n,j) \in \mathcal{E}} \!\!\! C_{n , j}^w \! - \!\!\!\! \sum_{j : (j,n) \in \mathcal{E}} \!\!\! C_{j , n}^w = 1, \; \forall w \in \mathcal{W}_n \setminus \{n\} \label{eq:steiner_root_flow} \\
		& \quad \sum_{j : (w,j) \in \mathcal{E}} \! \! \! C_{w , j}^w \! - \!\!\!\! \sum_{j : (j,w) \in \mathcal{E}} \!\!\! C_{j , w}^w = -1, \; \forall w \in \mathcal{W}_n \setminus \! \{n\} \label{eq:steiner_terminal_flow} \\
		& \quad \sum_{j : (i,j) \in \mathcal{E}} \!\!\!C_{i , j}^w \!-\!\!\!\! \sum_{j : (j,i) \in \mathcal{E}} \!\!\! C_{j , i}^w = 0, \notag \\
		& \qquad \qquad \qquad \qquad \quad  \forall w \in \mathcal{W}_n \setminus \{n\},\; \forall i \notin \{n,w\} \label{eq:steiner_intermediate_flow} \\
		& \quad \sum_{j : (n,j) \in \mathcal{E}} \!\!\! C^{\text{conn}}_{n , j} \!-\!\!\!\! \sum_{j : (j,n) \in \mathcal{E}} \!\!\!C^{\text{conn}}_{j , n} = \sum_{i \in \mathcal{N}} v_i - 1 \label{eq:conn_root_total_flow} \\
		& \quad \sum_{j : (j,i) \in \mathcal{E}} \!\!\! C^{\text{conn}}_{j , i} \! - \!\!\!\! \sum_{j : (i,j) \in \mathcal{E}} \!\!\! C^{\text{conn}}_{i , j} = v_i, \; \forall i \in \mathcal{N} \setminus \{n\} \label{eq:conn_node_flow_balance} \\
		& \quad C^{\text{conn}}_{i , j} \! \le \! |\mathcal{N}| \cdot h_{i , j}, \; C^{\text{conn}}_{j , i} \! \le \! |\mathcal{N}| \cdot h_{i , j}, \; \forall (i,j) \in \mathcal{E} \label{eq:conn_flow_capacity}
	\end{align}
\end{subequations}
where binary variable $h_{i,j}$ indicates whether edge $(i,j)$ is selected in the resulting tree, and $v_i$ indicates whether node $i$ is included. 
The auxiliary flow variable $C_{i,j}^w$ captures the unit-demand routing from the root node $n$ to each target node $w \in \mathcal{W}_n$ and contributes to the cost function in \eqref{eq:steiner_obj}. 
Constraints \eqref{eq:steiner_target_activated}--\eqref{eq:steiner_flow_edge_usage} enforce the tree structure and edge-node consistency.
Flow conservation for each demand is captured in \eqref{eq:steiner_root_flow}--\eqref{eq:steiner_intermediate_flow}.
To ensure global connectivity of the tree, an auxiliary flow variable $C^{\text{conn}}_{i,j}$ is introduced. 
This variable simulates pushing $\sum_{n \in \mathcal{N}} v_i -1$ units of flow from the root to all active nodes.
Constraint \eqref{eq:conn_root_total_flow} initiates the global flow from the root,
\eqref{eq:conn_node_flow_balance} ensures each activated node receives exactly one unit of flow,
and \eqref{eq:conn_flow_capacity} restricts global flow to edges that are less than predefined threshold $|\mathcal{G}|$, which guarantees that all activated nodes are connected within a single tree structure.

Problem~\eqref{eq:steiner_conn_ilp} is formulated as a mixed-integer linear programming (MILP) problem \cite{nemhauser1988integer} and can be solved by standard MILP solvers, such as PuLP and Gurobi. 
In general, exact MILP solving may incur exponential worst-case complexity, e.g., $\mathcal{O}(2^{|\mathcal{E}|+|\mathcal{N}|})$ \cite{schrijver1998theory}. 
However, this bound is overly pessimistic for the present formulation. 
The objective in \eqref{eq:steiner_obj} sums the routing cost of each terminal flow independently, without shared-capacity coupling across terminals. 
Therefore, the optimal value is simply $\sum_{w\in\mathcal{W}_n} \mathrm{dist}_{\mathcal{G}}(n,w)$, where $\mathrm{dist}_{\mathcal{G}}(n,w)$ denotes the minimum distance from the root $n$ to terminal $w$. 
This optimum can be attained by first computing a single-source shortest-path tree rooted at $n$ via one execution of Dijkstra's algorithm \cite{dijkstra2022note}, and then pruning all branches that do not contain any terminal node in $\mathcal{W}_n$, with resulting complexity $\mathcal{O}(|\mathcal{E}|+|\mathcal{N}|\log|\mathcal{N}|)$ \cite{fredman1987fibonacci}.
Hence, although the generic MILP formulation has an exponential worst-case complexity, the specific problem considered here admits an exact strongly polynomial-time solution by exploiting its structure.

\footnotesize
\bibliographystyle{IEEEtran}
\bibliography{IEEEabrv,./ref.bib}

\end{document}